\documentclass[11pt]{article}

\usepackage{times}
\usepackage{helvet}
\usepackage{courier}
\usepackage{graphicx} 
\usepackage{natbib}
\usepackage{algorithm}
\usepackage{algorithmic}
\usepackage{amsmath}
\usepackage{url}
\usepackage{amsmath}
\usepackage{amssymb}
\usepackage{mathrsfs}
\usepackage{array}
\usepackage{bm}
\usepackage{multirow}
\usepackage{hyperref}
\usepackage{tikz}
\usetikzlibrary{matrix}
\usepackage{comment}

\usepackage{subcaption}
\usepackage{wrapfig}

\usepackage{natbib}
\usepackage{etoolbox}
\usepackage{mathptmx}

\usepackage[margin=1in]{geometry}

\newcommand{\E}{\mathbb{E}}
\def\Cov{{\rm Cov}}
\def\N{{\rm N}}
\def\KL{{\rm KL}}
\def\P{P_{\rm data}}

\def\D{{\cal D}}
\def\G{{\cal G}}

\def\P{{q_{\rm data}}}

\def\E{{\rm E}}

\newcolumntype{L}[1]{>{\raggedright\arraybackslash}m{#1}}
\newcolumntype{R}[1]{>{\raggedleft\arraybackslash}m{#1}}
\newcolumntype{C}[1]{>{\centering\arraybackslash}m{#1}}

\begin{document}

\begin{center}
{\huge Representation Learning: A Statistical Perspective}\\
\vspace{3mm}

{\large Jianwen Xie$^1$, Ruiqi Gao$^2$, Erik Nijkamp$^2$, Song-Chun Zhu$^2$, and Ying Nian Wu$^2$}\\ \vspace{1mm}
{$^1$Hikvision Research Institute,  $^2$Department of Statistics, University California, Los Angeles } 
\end{center} \vspace{4mm}

\begin{center} 
\textbf{{\Large Abstract}}
\end{center} \vspace{-4mm}

Learning representations of data is an important problem in statistics and machine learning. While the origin of learning representations can be traced back to factor analysis and multidimensional scaling in statistics, it has become a central theme in deep learning with important applications in computer vision and computational neuroscience. In this article, we review recent advances in learning representations from a statistical perspective. In particular, we review the following two themes: (a) unsupervised learning of vector representations and (b) learning of both vector and matrix representations.

\vspace{2mm}

\section{Introduction}

Statistics is about understanding data. If the input data are complex, it is desirable to find representations for the data so that they become easier to understand and process.  In this article, we review learning representations of data with various models, including models with linear structures and models that are based on deep neural networks. 

\subsection{Prototypes of learning representations in statistics} 

Although representation learning is a central theme in deep learning, its essence can be traced back to familiar examples in statistics. 

\subsubsection{Factor analysis --- generative representation} 

One prototypical example of learning representation in statistics is {factor analysis} \citep{rubin1982algorithms}. Here, multivariate observations (e.g., test scores on different subjects) are explained by latent factors (e.g., verbal and analytical intelligence). Let $h$ be a $d$-dimensional hidden vector that consists of $d$ latent factors. Let $x$ be the observed $D$-dimensional vector. Usually $d < D$. Then, the model is of the form $x = W h + \epsilon$, where $W$ is the $D \times d$ loading matrix that transforms $h$ to $x$. It is assumed that $h \sim {\rm N}(0, I_d)$, where $I_d$ denotes the $d$-dimensional identity matrix, and $\epsilon \sim {\rm N}(0, \sigma^2 I_D)$, which is independent of $h$. This model can be learned by maximum likelihood via the expectation-maximization (EM) algorithm \citep{dempster1977maximum}, where the E step is based on the posterior distribution of $h$ given $x$.  

$h$ is said to be a {\em vector representation}, also called a {\em code} of $x$. The mapping from $h$ to $x$ is called a {\em decoder}, while the mapping from $x$ to $h$ is called an {\em encoder}, and both can be formally written as conditional distributions. While the decoder $p(x|h)$ and the prior $p(h)$ define a top-down {\em generative model}, the encoder $p(h|x)$ defines an {\em inference model}. 

Factor analysis is related to principal component analysis, where $W$ is obtained by the first $d$ eigenvectors of the covariance matrix $\Cov(x)$. The factor analysis model can be generalized to independent component analysis \citep{hyvarinen2004independent}, sparse coding \citep{olshausen1997sparse}, non-negative matrix factorization \citep{lee2001algorithms}, recommender systems \citep{koren2009matrix}, restricted Boltzmann machines \citep{hinton2012practical}, and so on, by modifying the prior distribution or prior assumption on $h$. If we generalize the linear mapping from $h$ to $x$ to a nonlinear mapping parameterized by a deep network \citep{lecun1998gradient, krizhevsky2012imagenet}, then the resulting model is commonly called {\em generator network}  \citep{goodfellow2014generative, KingmaCoRR13}.

Factor analysis is an example of ``{\em generative representation}'', where the hidden vector $h$ generates the observed vector $x$.

\subsubsection{Multidimensional scaling --- relative representation} 

The other prototypical example of learning representation in statistics is {multidimensional scaling} \citep{kruskal1964multidimensional}. Let $(x_i, i = 1, ..., n)$ be a set of $D$-dimensional observations. We want to represent them by a corresponding set of $d$-dimensional hidden vectors $(h_i, i = 1, ..., n)$, so that $(h_i)$ preserve the relations such as distances between $(x_i)$. For instance, we may find $(h_i)$ by minimizing $\sum_{i \neq j} (\|h_i - h_j\| - \|x_i - x_j\|)^2$, which enforces global {\em isometry}. 

Again $h$ is said to be a vector representation of $x$, which is also called an {\em embedding} of $x$. Unlike in factor analysis, there is no explicit mapping (encoding or decoding) between $h$ and $x$. 

Various modifications of multidimensional scaling focus on preserving local adjacency or neighborhood relations between $(x_i)$, such as spectral embedding \citep{bengio2004learning},   t-stochastic neighbor embedding (t-SNE) \citep{maaten2008visualizing}, and local linear embedding \citep{roweis2000nonlinear}. 

Multidimensional scaling is an example of  ``{\em relative representation}", where the hidden vectors $\{h_i\}$ are to preserve the relations between the observed vectors $\{x_i\}$. 

\subsubsection{Sliced inverse regression --- predictive representation} 

The third prototypical example of learning representation in statistics is sliced inverse regression \citep{li1991sliced}. It learns a nonlinear regression model from the training examples $\{(x_i, y_i)\}$, where $x_i$ is $D$-dimensional continuous predictor vector, and $y_i$ is one-dimensional continuous outcome. The sliced inverse regression model assumes a $d$-dimensional hidden vector $h_i = W x_i$, where $W$ is $d \times D$, so that $y_i = f(h_i, \epsilon_i)$ where $\epsilon_i$ are independent and identically distributed (i.i.d.) noise. 

Assume $(x_i, y_i) \sim p(x, y)$, $\E(x) = 0$, and $\Cov(x) = I_D$ under $p(x, y)$ (which can be achieved by standardizing $x$). Then under mild conditions, $W$ can be obtained by the top $d$ eigenvectors of $\Cov[\E(x|y)]$, where $\E(x|y)$ can be obtained by dividing the range of $y$ into slices, and $\E(x|y)$ is the inverse regression. $W$ can be obtained without knowledge of the nonlinear link function $f$. We may refer to $h = W x$ as encoding and $y = f(h, \epsilon)$ as decoding. 

Sliced inverse regression is an example of  ``{\em predictive representation}'', where the hidden vector $h_i$ contains all the information of $x_i$ for predicting $y_i$, i.e., $h_i$ is a sufficient summary of $x_i$ as far as predicting $y_i$ is concerned. 

\subsection{Unsupervised, supervised and reinforcement learning}

Sliced inverse regression is a supervised learning problem where for each input $x_i$, an output $y_i$ is given as supervision. Factor analysis and multidimensional scaling are unsupervised learning problems where only $x_i$ are observed without $y_i$. Learning representations is of fundamental importance for both supervised and unsupervised learning. In this article, we shall focus on unsupervised learning. 

Another learning problem that lies in between supervised and unsupervised learning is reinforcement learning \citep{sutton1998introduction}, where the input $x$ is the state, and the output $y$ is the action. In training, the optimal $y$ is not directly given, but a reward for an action is provided. For this problem, learning a good representation of state $x$ is important for  learning value and policy functions that are defined on the state. 

\subsection{Plan for the remainder of the article} 

Section \ref{sec:vector} presents vector representations based on linear models. We first describe a generalization of the factor analysis model in which the hidden vector is assumed to be sparse (or have independent components) in the generative representation scheme. We then explain continuous vector representations of discrete data, in predictive and relative representation schemes. Section \ref{sec:vector_matrix} presents the learning of both vector and matrix representations in a relative representation scheme. Section \ref{sec:generator} is about the learning of nonlinear vector representation based on the generator model, which generalizes linear mapping in the factor analysis model to nonlinear mapping parameterized by deep neural networks. Section \ref{sec:generator_joint} reviews the joint learning of generator model and various complementary models. Section \ref{sec:conditional} reviews the learning of the conditional generator model.

\section{Learning vector representations} 
\label{sec:vector}
In this section, we review learning vector representations of data using models that generalize the factor analysis model. 

\subsection{Sparse vector representation} \label{sect:s}

\begin{figure}[h]
	\centering	
		\begin{tabular}{ccc}
\includegraphics[height=.30\linewidth]{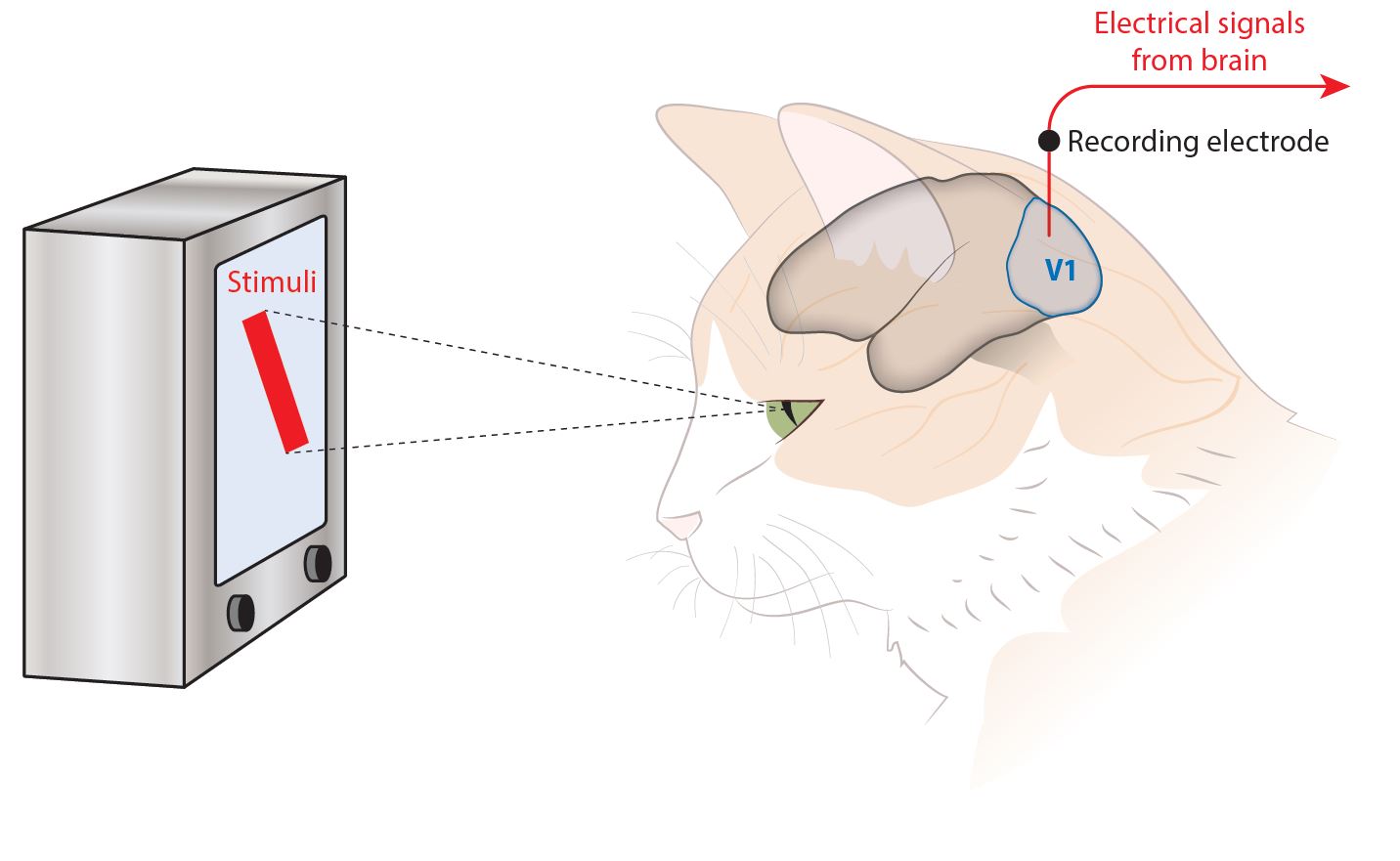}  \\
			\end{tabular}
\caption{\small Visual area of brain. Primary visual cortex, or V1, is the first step in representing retina image data. Cells in V1 respond to bars of different locations, orientations and sizes.  }	
	\label{fig:v1}
\end{figure}

David Hubel and Torsten Wiesel  earned  the Nobel Prize for Physiology or Medicine in 1981 for their discovery of simple and complex cells in the primary visual cortex or V1  \citep{hubel1959receptive}. They discovered that cells in V1 of the cat brain responded to bars of different locations, orientations and sizes, and each cell responded to the bar at a particular location, orientation and scale. See Figure \ref{fig:v1} for an illustration. Some V1 cells are called simple cells, which behave like linear wavelets. A mathematical model of a simple cell is the Gabor wavelet, which is a sine or cosine plane wave multiplied by an elongate Gaussian function.  

\cite{olshausen1997sparse} proposed a sparse coding model for the V1 simple cells by generalizing the factor analysis model. Recall that in factor analysis, 
\begin{eqnarray}
  x = W h + \epsilon = \sum_{k = 1}^{d} W_k h_k + \epsilon, 
\end{eqnarray}
where $W_k$ is the $k$-th column of $W$ and is of the same dimensionality as $x$, and $h_k$ is the $k$-th element of $h$. The above model expresses $x$ as a linear superposition of the basis vectors $W_k$, with coefficients $h_k$. 

Unlike in factor analysis, in the sparse coding model, the dimensionality $d$ of $h$ is assumed to be larger than the dimensionality $D$ of $x$ (i.e., $d > D$). However, $h$ is assumed to be a sparse vector, i.e., for each $x$, only a small number $d_0$ ($d_0 < D < d$) of $h_k$ are non-zero or significantly different from zero. For different $x$, the non-zero elements of $h$ can be different. Thus unlike principal component analysis, sparse coding leads to adaptive dimension reduction. $W = (W_k, k = 1, ..., d)$ is sometimes called a ``dictionary'', from which a small number of ``words'' are chosen to describe $x$. $h$ is called a {\em sparse code} of $x$. 

\begin{figure}[h]
	\centering	
		\begin{tabular}{c}
	\includegraphics[height=.3\linewidth]{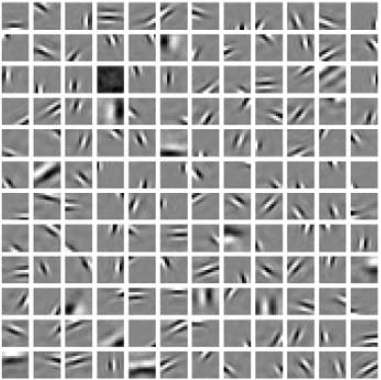} 
			\end{tabular}
		\caption{\small Olshausen-Field sparse coding model. The plot displays the 144 learned basis vectors, each displayed as an image patch (ordering of the patches carries no meaning).  These basis vectors represent local image structures such as edges and bars. The training data were obtained by extracting $12 \times 12$ image patches at random from ten $512 \times 512$  images of natural scenes (trees, rocks, mountains etc.). }	
	\label{fig:sparse}
\end{figure}

The training data are in the form of image patches sampled from natural images, $\{x_i, i = 1, ..., n\}$, where each $x_i$ is a training example image patch. Each $x_i$ is represented by an $h_i = (h_{ik}, k = 1, ..., d)$, but all the examples share the same $W$, where each $W_k$ has the same dimensionality as $x_i$, so that $x_i = W h_i + \epsilon_i = \sum_{k=1}^{d} W_k h_{ik} + \epsilon_i$. The learning of $W$ can be accomplished by minimizing the following objective function 
\begin{eqnarray}
    L(W, \{h_i\}) = \frac{1}{n} \sum_{i=1}^{n} \left[ \|x_i - W h_i\|^2  + \sum_{k=1}^{d} \rho(h_{ik})\right],
 \end{eqnarray}
where $\rho(h_{ik})$ is a sparsity-inducing term, e.g., $\rho(r_{ik}) = |r_{ik}|$, which leads to the Lasso estimator  \citep{tibshirani1996regression} of $h_i$. The minimization can be accomplished by alternating gradient descent over $W$ and $\{h_i\}$. Figure~\ref{fig:sparse} displays the learned $(W_k)$, where each $W_k$ is displayed as an image patch of the same size as $x_i$. The basis vectors $(W_k)$ represent local image structures such as bars and edges. 

Given $W$, the inference of $h_i$ from each $x_i$ can be accomplished by the Lasso, where $(W_k)$ serve as variables or regressors. Compared to the Lasso, the sparse coding has an added layer of depth in that $W$ (i.e., the regressors) is to be learned from the training data. The sparse coding model has had a profound impact on computational neuroscience and applied harmonic analysis,  in addition to machine learning. 

A related model is {independent component analysis}  \citep{bell1997independent, hyvarinen2004independent}, which assumes that $D = d$, $\epsilon = 0$, and $h_k$ are independent. It assumes an invertible transformation $x = W h$, and $h = W^{-1} x$, so that the distribution of $x$ can be obtained in closed form from the prior distribution of $h$: $p(x) = p_0(W^{-1}x) |W|^{-1}$, where $p_0(h)$ is the prior distribution of $h$, and $|W|$ is the absolute value of the determinant of $W$.  

Other related models include non-negative matrix factorization \citep{lee2001algorithms}, which assumes $h_k \geq 0$, and restricted Boltzmann machines \citep{hinton2012practical}, which assume a binary $h$ and a joint distribution $p(x, h) \propto \exp(- x^\top W h)$ (where we omit bias terms for simplicity), which is an energy-based model on $(x, h)$ with pairwise potentials defined on $(x, h)$. For this model, both the decoder $p(x|h)$ and the encoder $p(h|x)$ are in closed form. But the prior distribution $p(h)$ is not in closed form.

\subsection{Continuous vector representation of discrete or symbolic input} 

The vector representation $h$ of the original input $x$ can be considered a dimension reduction of $x$,  or visualization of $x$ if $h$ is 2D ($d = 2$). The input $x$ is usually continuous. 

The input $x$ can also be discrete, like a word in the dictionary. In that case, $x$ can be expressed as a one-hot vector. Let $D$ be the number of words in the dictionary. If $x$ is the $j$-th word in the dictionary, then $x$ is a $D$-dimensional vector so that the $j$-th element of $x$ is 1 and all the other elements are zeros. We represent $x$ with a $d$-dimensional continuous hidden vector $h$. We can write $h = W x$, where $W$ is a $d \times D$ dimensional encoding matrix, so that the $j$-th word is represented by the $j$-th column of the encoding matrix $W$. $h$ is called a semantic embedding or word2vec \citep{mikolov2013efficient, pennington2014glove}. In \cite{mikolov2013efficient}, $h$ is learned to predict nearby words, i.e., it is a predictive representation. Specifically, for a particular word $y$, again expressed as a one-hot vector,  in the context of word $x$ in a random sentence, we predict this word $y$ based on the decoded vector $\tilde{W}^\top h$,  where $\tilde{W}$ is the $d \times D$ decoding matrix of the same dimensionality as the encoding matrix $W$,  so that $p(y) \propto \exp(y^\top \tilde{W}^\top h)$. More specifically, let $Q_{ij}$ be the probability that word $j$ is within the context of word $i$, then $Q_{ij} = \exp(\langle W_i, \tilde{W}_j\rangle)/\sum_j \exp(\langle W_i, \tilde{W}_j\rangle)$, the so-called soft-max classifier, where $W_i$ is the $i$-th column of $W$, i.e., the vector representation of word $i$ in the encoding pass, and $\tilde{W}_j$ is the $j$-th column of $\tilde{W}$, i.e., the vector representation of word $j$ in the decoding pass. 

In \cite{pennington2014glove}, $h = Wx$ is learned as a relative representation so that for two words $i$ and $j$, $\log Q_{ij} = \langle W_i, \tilde{W}_j\rangle + b_i  + \tilde{b}_j$, where $b_i$ and $\tilde{b}_j$ are bias terms. 

The above form is similar to matrix factorization in recommender systems \citep{koren2009matrix}. Let $X_{ij}$ be  the rating of user $i$ on item $j$, and then the model is  $X_{ij} = \langle W_i, \tilde{W}_j\rangle + b_i  + \tilde{b}_j$, where $W_i$ is the vector representation of user $i$, $\tilde{W}_j$ is the vector representation of item $j$, and $b_i$ and $\tilde{b}_j$ are the bias terms. The elements of the $d$-dimensional vector $W_i$ can be interpreted as the desires of user $i$ in various aspects, and  the elements of the $d$-dimensional $\tilde{W}_j$ can be interpreted as the desirabilities of item $j$ in the corresponding aspects. In terms of matrix, let $X$ be the $n \times D$ matrix of ratings where $n$ is the number of users and $D$ is the number of items. Then $X  = W^\top \tilde{W}$, where $W$ is the $d \times n$ matrix whose $i$-th column is $W_i$, and $\tilde{W}$ is the $d \times D$ matrix whose $j$-th column is $\tilde{W}_j$. 

 For a discrete $x$ such as a word, the vector representation $h$ is continuous, dense, and distributed, where each component of $h$ captures a partial semantic meaning of $x$. Such dense vector representations have revolutionized natural language processing in recent years, and they are at the foundation of recent natural language models \citep{vaswani2017attention, radford2018improving, devlin2018bert}. 

Vector representations have also been applied to encode the nodes in graphs \citep{hamilton2017representation}, which can be conveniently used for subsequent analysis  \citep{kipf2016semi}. 

In \cite{gomez2018automatic}, each molecular compound, which is a graph structure, is represented by a continuous vector, which can be used to learn to predict the chemical activity of the compound. One can also optimize the activity by maximizing over the continuous vector using gradient-based method, and the optimized vector can then generate the corresponding compound. Such continuous representation is much more convenient to operate on than the original discrete input. 

\section{Learning both vector and matrix representations}  
\label{sec:vector_matrix}

This section reviews recent work on learning  models based on vector and matrix representations. The representations are of a relative nature, similar to multidimensional scaling. The matrices represent the relations between the vectors, and can be part of a relative representation. An early example is that of \cite{paccanaro2001learning}. 

 In computational neuroscience, the vector representations can be interpreted as neuron activities, and the matrix representations can be stored in the synaptic connections. The vector representations are like ``nouns'', while the matrix representations are like ``verbs'' that transform the ``nouns''. 

Matrix representations of groups  underlie much of modern mathematics \citep{dornhoff1972group} and hold the key to modern physics \citep{zee2016group}. 

\subsection{Learning grid cells} 

You may imagine moving in your living room at night in the dark. Based purely on the movements or self-motion, you know your current position by summing up the displacements. The grid cells in our brain accomplish this computation, albeit in a very sophisticated manner. 

\subsubsection{Hexagon patterns} 

\begin{figure}[h]
\centering	
\begin{tabular}{cccc}
\includegraphics[height=.15\linewidth]{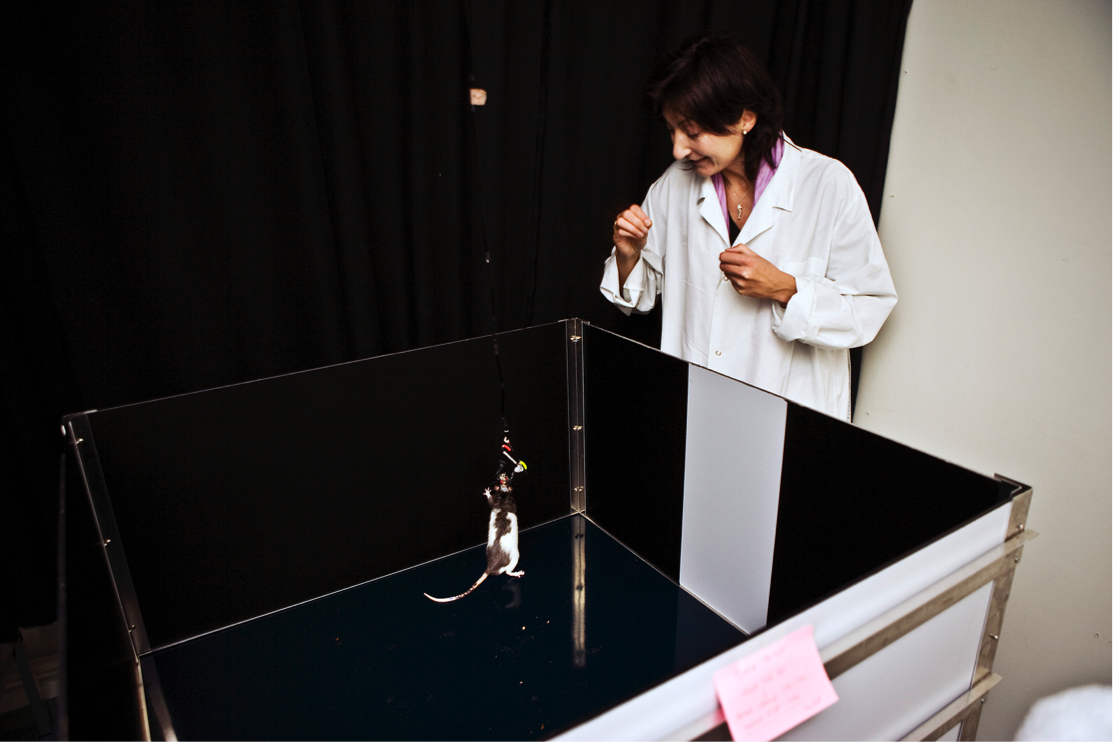}  &
\includegraphics[height=.15\linewidth]{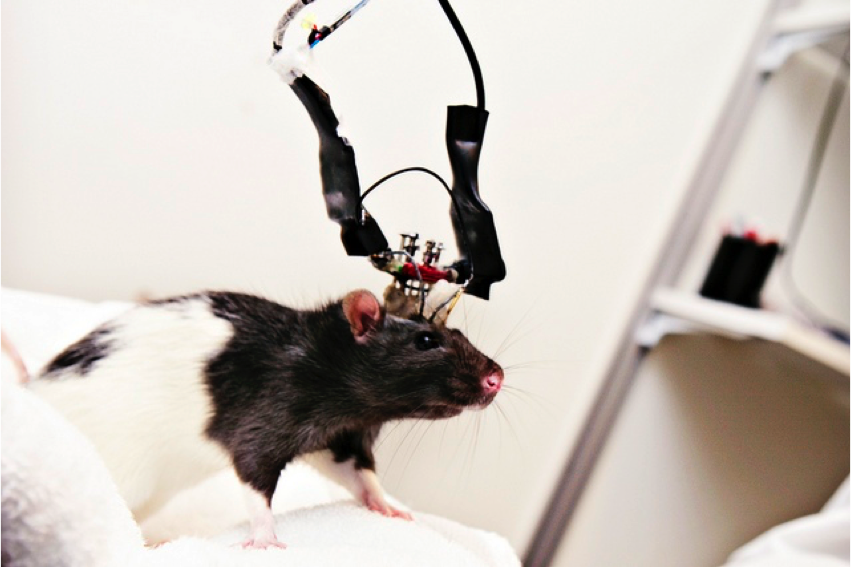} &
\includegraphics[height=.15\linewidth]{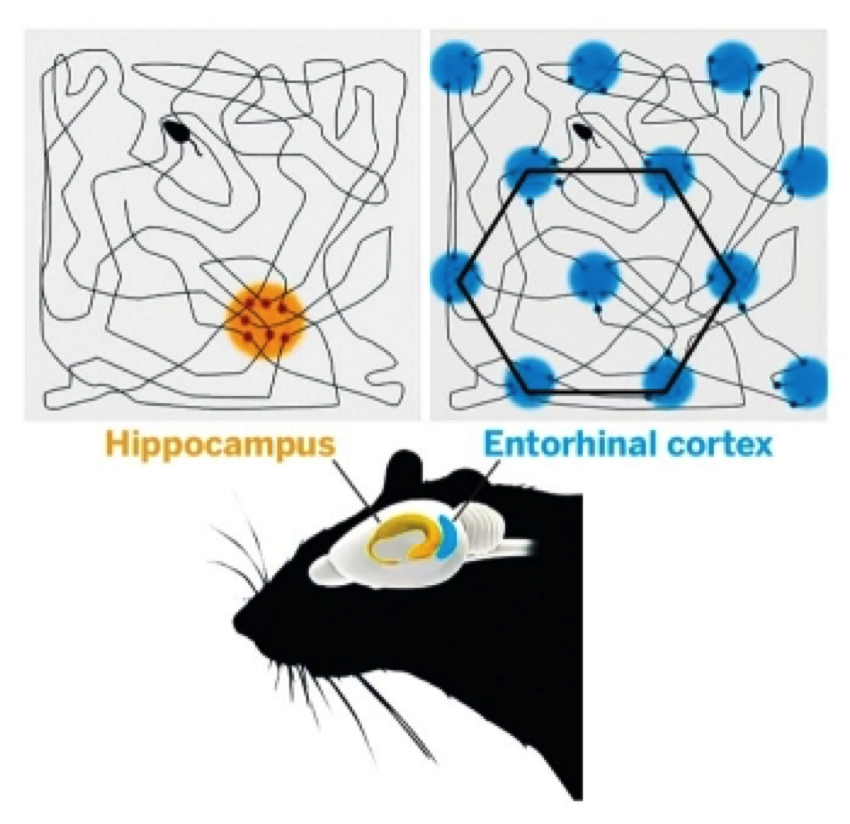} &
\includegraphics[height=.15\linewidth]{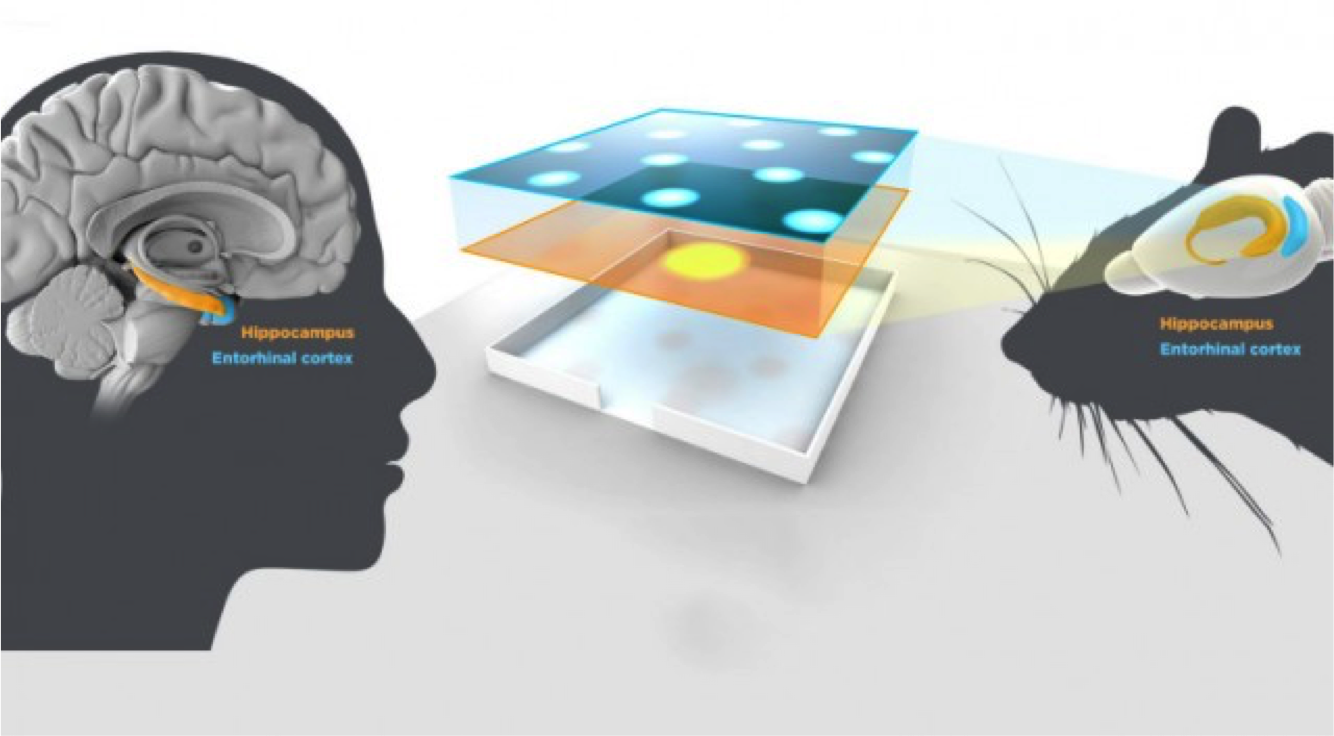} \\
	(a) & (b) & (c) &(d) 
	\end{tabular}
		\caption{\small Place cells and grid cells. (a) The rat is moving within a square region. (b) The activity of a neuron is recorded. (c) When the rat moves around (the curve is the trajectory), each place cell fires at a particular location, but each grid cell fires at multiple locations that form a hexagon grid. (d) The place cells and grid cells exist in the brains of both rat and human. }	
	\label{fig:g1}
\end{figure}

Figure \ref{fig:g1}a depicts Dr. May-Britt Moser who, together with Dr. Edvard Moser, won the 2014 Nobel Prize for Physiology or Medicine, for the discovery of the grid cells \citep{hafting2005microstructure}. Their thesis advisor, Dr. John O'Keefe, shared the prize for his discovery of place cells \citep{o1979review}. Both place and grid cells are used for navigation. The discoveries of these cells were made by recording the activities of the neurons of a rat when it moves within a square region (Figure \ref{fig:g1}b). Some neurons in the hippocampus area are place cells. Each place cell fires when the rat moves to a particular location, and different place cells fire at different locations; the whole collection of place cells covers the whole square region. The discovery of grid cells, found in the entorhinal cortex, was much more surprising and unexpected. Each grid cell fires at multiple locations, and these locations form a regular hexagonal grid (Figure \ref{fig:g1}c). The grid cells have been identified across many mammalian species, including human (Figure \ref{fig:g1}d).

\subsubsection{A simple addition problem} 

There are two problems in navigation. One is the path integral. Imagine you walk in your living room at night. If you know the position of your starting point, then by summing over your displacements over time, you can calculate where you are at any time. The other problem is path planning. Suppose you want to go to a target position such as the light switch, which is a position that you know, so you plan a sequence of displacements that will lead you from the starting point to the target.  

More specifically, consider an agent (e.g., a rat or a human) navigating within a domain $D = [0, 1] \times [0, 1]$. We can discretize $D$ into an $N \times N$ lattice. Let $x = (x^{(1)}, x^{(2)}) \in D$ be the self-position of the agent. Let $\Delta x = (\Delta x^{(1)}, \Delta x^{(2)})$ be the displacement or self-motion of the agent at a certain time. The path integral problem is such that, given the starting point $x_0$ and the sequence of self-displacements $(\Delta x_t, t = 1, ..., T)$, we want to calculate the positions over time with $x_{t} = x_{t-1} + \Delta x_t$ for $t = 1, ..., T$. The path planning problem is, given the starting position $x$ and the target position $y$, to plan a sequence of displacements $(\Delta x_t, t = 1, ..., T)$ such that $x_0 = x$ and $x_T = y$. 

Both problems appear to be quite simple, especially path integral, which is merely an addition problem. But the brain uses a system of grid cells to solve this problem. What is the purpose of this system and how does the system work? Why the hexagon patterns?

\subsubsection{A representational scheme} 

Recently  \cite{gao2018gridcell_learning} proposed an explanation of grid cells as a representational system. The basic idea is that the grid cells form a $d$-dimensional vector representation of the 2D position. Specifically, we represent any 2D position $x \in D$ by a $d$-dimensional vector $h(x)$. Suppose at a position $x$, the self-motion or displacement is $\Delta x$, so that the agent moves to $x + \Delta x$ after one step. We assume the following motion model: 
\begin{eqnarray}
     h(x+\Delta x) = M(\Delta x) h(x), \label{eq:motion}
\label{eqn: motion}
\end{eqnarray}      
where $M(\Delta x)$ is a $d \times d$ matrix that depends on $\Delta x$. While $h(x)$ is the vector representation of the self-position $x$, $M(\Delta x)$ is the matrix representation of the self-motion $\Delta x$.  As we show below, $\|h(x)\| = 1$ for all $x$, thus $M(\Delta x)$ is a rotation matrix, and the self-motion in 2D is represented by a rotation in the $d$-dimensional sphere. We can illustrate the motion model by the following diagram: 
\begin{eqnarray}
\begin{array}[c]{cccc}
& x_t & \stackrel{ + \Delta x}{\xrightarrow{\hspace*{1cm}}}& x_{t + 1} \\
& \downarrow& \downarrow &\downarrow\\
& h(x_t) &\stackrel{ M(\Delta x) \times }{\xrightarrow{\hspace*{1cm}}} & h(x_{t+1}) 
\end{array}  \label{eq:diagram}
\end{eqnarray}
Both $h(x)$ and $M(\Delta x)$ are to be learned. 

\begin{figure}[h]
\centering	
\begin{tabular}{cc}
\includegraphics[height=.11\linewidth]{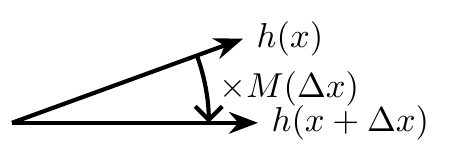}  &
\includegraphics[height=.11\linewidth]{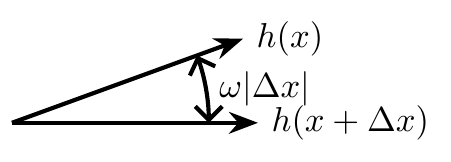} \\
(a) Vector-matrix multiplication & (b) Magnified local isometry 
\end{tabular}
\caption{\small Grid cells form a high-dimensional vector representation of 2D self-position. Two sub-models: (a) Local motion is modeled by vector-matrix multiplication. (b) Angle between two nearby vectors magnifies the Euclidean distance. $x$ is a 2D position, $\Delta x$ is a 2D self-motion, $h(x)$ is a high-dimentional vector representation of $x$, and $M(\Delta x)$ is a matrix representation of $\Delta x$. $\omega$ is a magnifying factor.}
\label{fig:vector_plot}
\end{figure}

\cite{gao2018gridcell_learning} proposed that the brain uses the above representational scheme to carry out the simple addition calculation (illustrated in Figure \ref{fig:vector_plot}a; see also \citep{paccanaro2001learning} for an earlier treatment of the addition problem). 

\subsubsection{Error correction} 

In data visualization such as t-SNE \citep{maaten2008visualizing}, we represent high-dimensional data by 2D points. In grid cells, we do the opposite, representing 2D coordinates by high-dimensional vectors. Why does the brain bother with a high-dimensional representation of 2D coordinates? The answer lies in error correction. The neurons are intrinsically noisy. For a noisy observation of $h(x)$, by projecting it onto the sub-manifold of $(h(x), x \in [0, 1]^2)$, we can eliminate most of the noise.  

In order to reduce the noise, we can use a high-dimensional $h$ to record multiple noisy copes of $x$; then, a simple averaging will reduce the variance of noise. Apparently the brain goes much further than that. It represents 2D $x$ by a high-dimensional $h$, so that the angle between $h(x)$ and $h(x+\Delta x)$ is $\omega |\Delta x|$ for $\omega \gg 1$. This makes the system even more robust to noise, because $h(x)$ and $h(x+\Delta x)$ are very far apart when $\omega \gg 1$. 

More specifically, we assume a magnified  local isometry  model:   
\begin{eqnarray}
     \langle h(x), h(x+\Delta x)\rangle = 1 - \alpha |\Delta x|^2, \label{eq:localization0} 
 \end{eqnarray}
 which is a second-order Taylor expansion of a function of $|\Delta x|$ whose maximum is 1 at $|\Delta x| = 0$. For $\Delta x = 0$,  we have $\|h(x)\|^2 = 1$ for all $x$. Let $\Delta \theta$ be the angle between $h(x)$ and $h(x+\Delta x)$, and then,  $\langle h(x), h(x+\Delta x)\rangle = \cos(\Delta \theta) \approx 1  - \Delta \theta^2/2$ for small $\Delta \theta$. Thus $\Delta \theta$ is proportional to $|\Delta x|$, i.e., $\Delta \theta = \omega |\Delta x|$, where $\omega = \sqrt{2 \alpha} \gg1$ (See Figure \ref{fig:vector_plot}b). 

 \cite{gao2018gridcell_learning}  showed that even if they randomly shut down (i.e., set to zero) $70\%$ of the neurons in each step, their learned system can still perform a path integral accurately. Such dropout error could occur due to internal noises and asynchrony of neuron activities, as well as aging and diseases like Alzheimer's. 

 \subsubsection{Emergence of hexagon patterns} 
 
 For a fixed $\alpha$, we can learn $(h(x), \forall x)$ and  $M(\Delta x)$ by minimizing the least-squares loss: 
\begin{eqnarray}
  \E_{x, \Delta x} [ \|h(x+\Delta x) - M(\Delta x) h(x)\|^2] + \lambda \E_{x, \Delta x}[(\langle h(x), h(x+\Delta x)\rangle - (1- \alpha |\Delta x|^2))^2]. 
\end{eqnarray}
The above loss function can be minimized by stochastic gradient descent, where for stochastic approximation of the expectations, we randomly sample $(x, \Delta x)$ uniformly where $x \in [0, 1]^2$ and $\Delta x$ is within a limited range.

 \begin{figure}[h]
  \begin{center}
  	\begin{minipage}[b]{.455\textwidth}
\centering
\begin{tabular}{C{.15\textwidth}C{.75\textwidth}}
 {\tiny $\alpha = 18$} & \includegraphics[width=.75\textwidth]{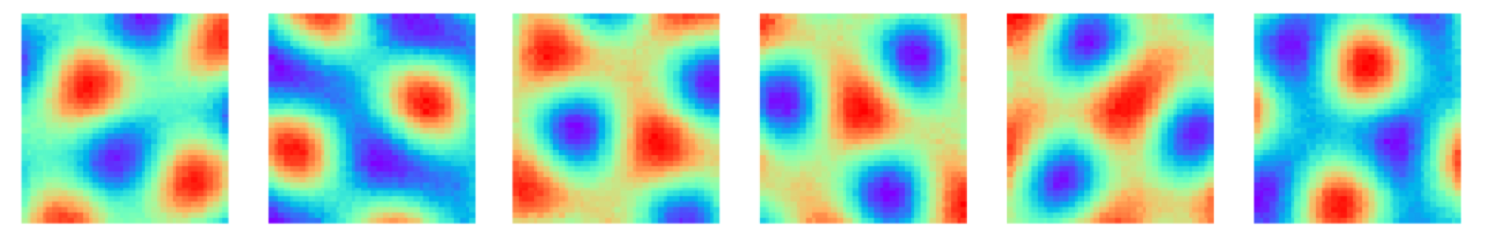} \\
 
 {\tiny $\alpha = 36$}  & \includegraphics[width=.75\textwidth]{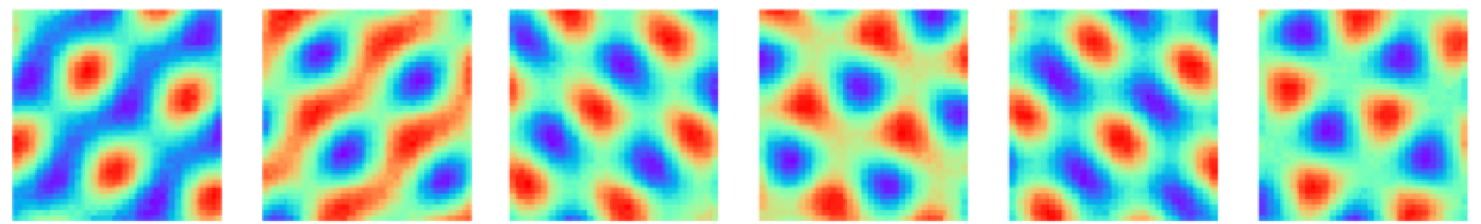} \\ 
 
  {\tiny  $\alpha = 72$} & \includegraphics[width=.75\textwidth]{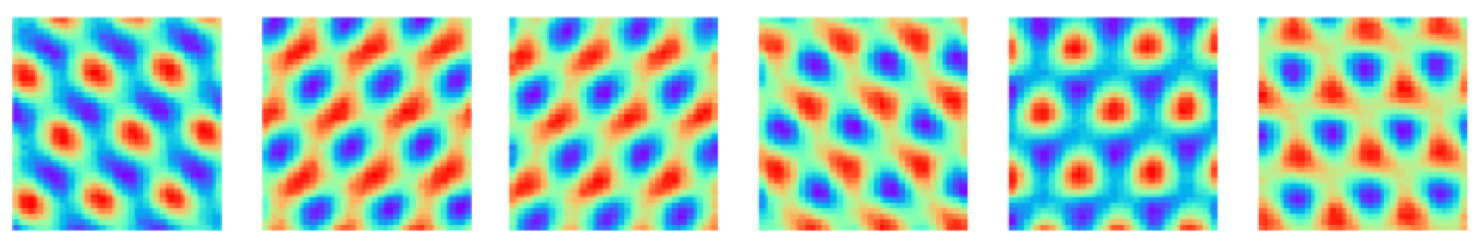}\\

  {\tiny $\alpha = 108$}  & \includegraphics[width=.75\textwidth]{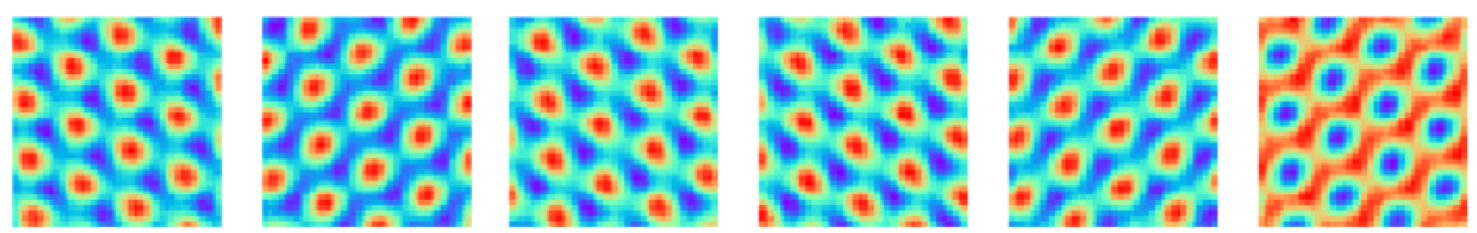} \\
	  {\tiny $\alpha = 144$} & \includegraphics[width=.75\textwidth]{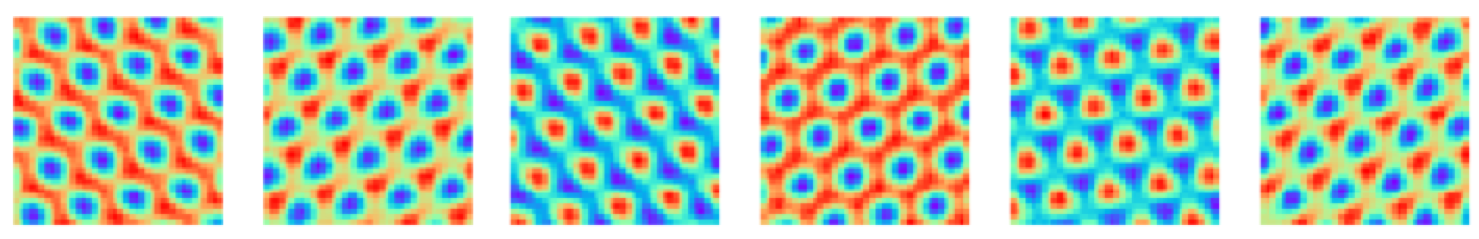} \\
	 {\tiny $\alpha = 180$} & \includegraphics[width=.75\textwidth]{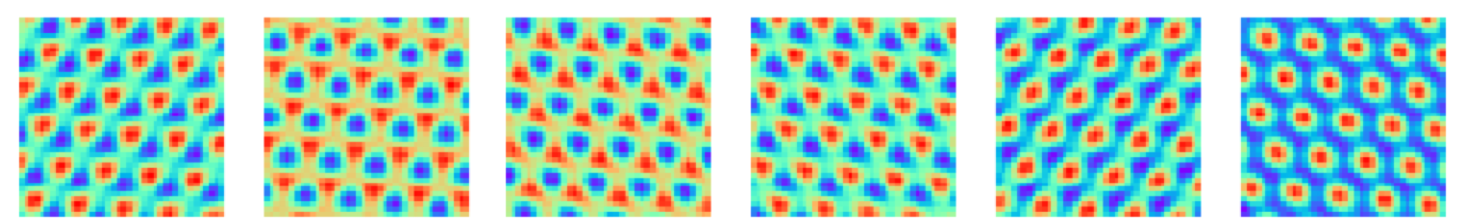} 
\end{tabular} 
{\small (a) Learned $h(x)$ with $6$ units. }
\end{minipage}
\qquad
\begin{minipage}[b]{.36\textwidth}
	\centering
	\includegraphics[width=\textwidth]{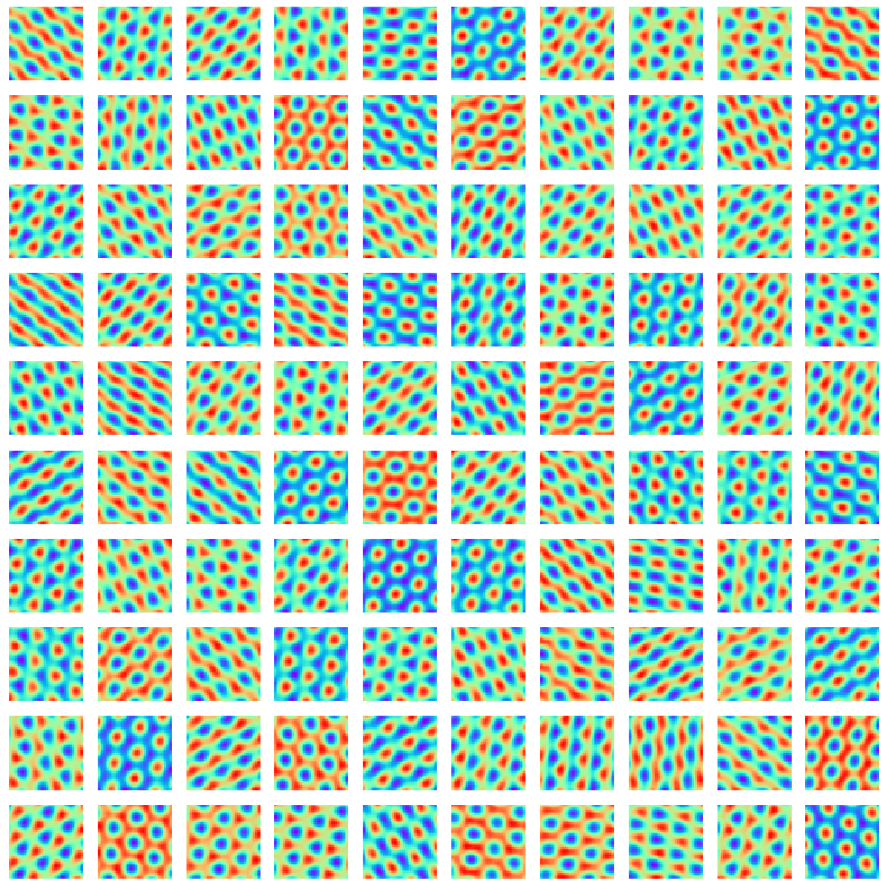}
	{\small (b) Learned $h(x)$ with $100$ units}
\end{minipage}
  \end{center}
\caption{\small Learned grid cells. (a)  Each row shows a component or a unit of $h(x)$ with a certain metric parameter $\alpha$, where the number of units $d$ is 6. (b) Learned units where the number of units $d$ is $100$ and $\alpha = 72$.}
\label{fig:g2}
\end{figure}

The experiments of  \cite{gao2018gridcell_learning} show that as long as the dimension of $h(x)$, $d \geq 6$, then the learning algorithm always learns the hexagon grid pattern for each element of $h(x)$. Even if $d = 100$, the algorithm still learns regular hexagon patterns. In Figure \ref{fig:g2}a, each row displays the learned $h(x)$ for a given value of $\alpha$, where $d = 6$.  Figure \ref{fig:g2}b shows the learned $h(x)$ for $\alpha = 72$, where $d = 100$. If $d<6$, the algorithm tends to learn square grid patterns. 

Thus if we move from $x$ to $x+\Delta x$, the corresponding $h$ will be  rotated by a matrix $M(\Delta x)$, and $h$ will rotate at a much faster speed $\omega |\Delta x|$.  As a result, it will quickly rotate back to itself, causing the periodic grid patterns, which in turn causes the global ambiguities in position, because the same $h$ may correspond to multiple positions. One could say that the grid patterns are almost an unwanted consequence of error correction. To resolve the ambiguities, \cite{gao2018gridcell_learning} combined multiple blocks of grid cells to determine the position uniquely, and for each block, the magnifying parameter $\alpha$ can be learned automatically.

\subsection{Vector representation of state and matrix representation of motion or action} 

We may generalize the model in the previous subsection into a more general model for dynamic systems, where we represent the state by a vector and the change of the state caused by motion or action by a matrix. For example, \cite{gao2018V1learning} recently proposed a model of V1 simple cells that is different from the sparse coding model \citep{olshausen1997sparse} and the independent component analysis model \citep{bell1997independent} reviewed in Section \ref{sect:s}. \cite{gao2018V1learning} proposes that a direct purpose of V1 cells is to perceive the displacements of pixels over time, where the displacements of pixels are caused by the relative motion between the agent (a rat or a human) and the surrounding 3D environment. Specifically, \cite{gao2018V1learning} represents the local image contents by vectors, and the local displacements of pixels by matrices, so that when the pixels undergo displacements, the vectors are rotated by the matrices representing the displacements. After learning this representational system, the agent will be able to sense the displacements of pixels based on the rotations of the vectors. 

More generally, for a video sequence, we can represent the image frames by vectors, and the motions or actions of the agent or the objects in the image by matrices. This will enable the agent to perceive the objects and their motions and actions, while the agent is moving or taking actions. 

In terms of neuroscience, the vectors correspond to the activities of neurons, and the matrices correspond to the synaptic connections.  Interestingly, such a representational scheme appears to occur in nature. In quantum theory, the states are represented by vectors in a Hilbert space, and the changes of the states are represented by matrices or operators \citep{zee2016group}. Similar to the creation and annihilation operators in quantum field theory, the matrix representations in vision may also account for discrete events such as the appearance and disappearance of objects. Perhaps the brain speaks the same mathematical language as nature.

\section{Learning nonlinear vector representation by generator network}  
\label{sec:generator}

This section reviews the generator network that is a generalization of factor analysis where the mapping from the latent factors to the signal is parameterized by a deep network. We also discuss the maximum likelihood learning algorithm that learns various generator models. 

\subsection{Deep neural networks} 

The models reviewed so far are based on linear structures. They can be generalized to nonlinear transformations, such as deep neural networks  \citep{lecun1998gradient, krizhevsky2012imagenet}, which are compositions of multiple layers of linear transformations and coordinate-wise nonlinear link functions.  

Specifically, consider a nonlinear transformation $f(x)$ that can be decomposed recursively as $s_{l} = W_l h_{l-1} + b_l$, and $h_l = r_l(s_l)$, for $l = 1, ..., L$, with $f(x) = h_L$ and $h_0 = x$. $W_l$ is a weight matrix at layer $l$, and $b_l$ is the bias vector at layer $l$. Both $s_l$ and $h_l$ are vectors of the same dimensionality, and $r_l$ is a one-dimensional nonlinear link function, the rectification function, that is applied coordinate-wise. $f(x)$ is a recursive composition of generalized linear model (GLM) structures. 

Modern deep networks usually use $r_l(s) = \max(0, s)$, the rectified linear unit or ReLU. For such nonlinear link functions, $f(x)$ is a multivariate linear spline where the linear pieces are recursively partitioned. This is similar to but more general than the recursive partitions in classification and regression trees (CART) \citep{breiman2017classification} and multivariate adaptive regression splines (MARS) \citep{friedman1991multivariate}. 

In computational neuroscience, each element or unit in $h_l$ can be interpreted as a {\em neuron} or a {\em cell}, whose value can be related to the firing rate. Sometimes $h_l$ is colloquially called a {\em thought vector}. 

There are two special classes of neural networks. One consists of convolutional neural networks  \citep{lecun1998gradient, krizhevsky2012imagenet}, which are commonly applied to images,  where the same linear transformations are applied around each pixel locally. The other class consists of recurrent neural networks \citep{hochreiter1997long}, which are commonly applied to sequence data such as speech and natural language. 

Neural networks are commonly used in supervised learning and reinforcement learning, where $h_l$ at multiple layers can be considered predictive representations. They are also useful for unsupervised learning of generative models, as we review in the next subsection, where $h_l$ at multiple layers can be considered generative representations. 

\subsubsection{Nonlinear generalization of logistic regression} \label{sect:l}

For the deep network reviewed in the previous subsection, let $\alpha = (W_l, b_l, l = 1, ..., L)$ collect all the weight and bias parameters, and let $f_\alpha(x)$ be the resulting nonlinear transformation. 

We can generalize the logistic regression model to 
\begin{eqnarray}
   {\text P}(y = 1|x) = D(x) = \frac{1}{1+\exp(-f_\alpha(x))}. \label{eq:discriminator}
\end{eqnarray}
The model is also called a discriminator network, and $h_l$ at different layers can be considered predictive representations of $x$. 

\subsubsection{Nonlinear generalization of the exponential family model} \label{sect:e}

We can also generalize the exponential family model to
\begin{eqnarray} 
   \pi_\alpha(x) = \frac{1}{Z(\alpha)} \exp(f_\alpha(x)) \rho(x), \label{eq:exp}
\end{eqnarray} 
where $\rho(x)$ is a reference measure such as the uniform distribution and $Z(\alpha)$ is the normalizing constant. This model is also called the energy-based model or Gibbs distribution. 

The connection between the two models are as follows. Suppose $\rho(x)$ is the distribution of negative examples, i.e., ${\text P}(x|y = 0) = \rho(x)$, and $\pi_\alpha(x)$ is the distribution of positive examples, i.e., ${\text P}(x|y = 1) = \pi_\alpha(x)$. Suppose there are equal numbers of positive and negative examples; then, according to the Bayes rule, ${\text P}(y = 1|x)$ is given by Equation \ref{eq:discriminator}. 

Later in the article, we make use of the above two models as the complementary models to the generator model, which we review next. 

\subsection{Nonlinear generalization of factor analysis and maximum likelihood learning} 

While sparse coding and independent component analysis etc. generalize the prior assumption on the hidden vector $h$ in factor analysis, the generator model generalizes the mapping from the hidden vector $h$ to the input $x$, i.e., 
\begin{eqnarray} 
h \sim {\rm N}(0, I_d), \; x = g_\theta(h) + \epsilon,
\end{eqnarray} 
where $g$ is parameterized by a deep network, similar to $f$ in the previous subsection, i.e., $s_{l} = W_l h_{l+1} + b_l$, and $h_l = r_l(s_l)$, for $l = L-1, ..., 0$, with $h_L = h$, and $x = h_0$. $W_l$ is a weight matrix at layer $l$, and $b_l$ is the bias vector at layer $l$.  $\theta$ collects all the weight and bias parameters at all the layers, and $\epsilon \sim {\rm N}(0, \sigma^2 I_D)$ is the residual noise image that is independent of $h$.  

While $f$ in the previous subsection is a bottom-up network in the sense that it defines $h_0 = x \rightarrow h_1 \rightarrow ... \rightarrow h_L$, $g$ in this subsection is a top-down network in the sense that it defines $h_L = h \rightarrow h_{L-1} \rightarrow ... \rightarrow h_0 = x$. 

As in factor analysis, the model can be learned by maximum likelihood. We can write the prior distribution as $h \sim p(h)$, where $p(h)$ is the density of ${\rm N}(0, I_d)$. The conditional distribution of $x$ given $h$ is $p_\theta(x|h)$, which is the density of ${\rm N}(g_\theta(h), \sigma^2 I_D)$. The joint distribution or the complete-data model is $p_\theta(h, x) = p(h) p_\theta(x|h)$. The marginal distribution, or the observed-data model, is $p_\theta(x) = \int p_\theta(h, x) dh$. The posterior distribution of $h$ given $x$ is $p_\theta(h|x) = p_\theta(h, x)/p_\theta(x)$. Unlike in factor analysis, the marginal $p_\theta(x)$ and the conditional $p_\theta(h|x)$ are not in closed form. 

Let $q_{\rm data}$ be the distribution that generates the observed examples $x_i, i = 1, ..., n$. For large $n$, the maximum likelihood estimation of $\theta$ is to minimize the Kullback-Leibler (KL) divergence ${\rm KL}(q_{\rm data}\| p_\theta)$ over $\theta$, where the KL divergence is defined as ${\rm KL}(q|p) = \E_q[\log (q(x)/p(x))]$. In practice, the expectation with respect to $q_{\rm data}$ is approximated by the average over the observed examples. The gradient of the log-likelihood can be computed based on
\begin{eqnarray} 
    - \frac{\partial}{\partial \theta} {\rm KL}(q_{\rm data}(x)\|p_\theta(x)) = \E_{q_{\rm data}(x) p_\theta(h|x)} \left[\frac{\partial}{\partial \theta} \log p_\theta(h, x)\right]. 
    \label{eq:abp}
\end{eqnarray} 
The expectation with respect to the posterior distribution $p_\theta(h|x)$ can be approximated via Markov chain Monte Carlo (MCMC) sampling of $p_\theta(h|x)$,  such as Langevin dynamics or Hamiltonian Monte Carlo (HMC) \citep{neal2011mcmc}.  It can be efficiently implemented by gradient computation via back-propagation.

\begin{figure}[h]
\begin{center}
 \includegraphics[width=.6\columnwidth]{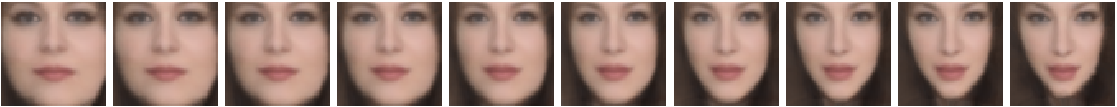}\\
 \includegraphics[width=.6\columnwidth]{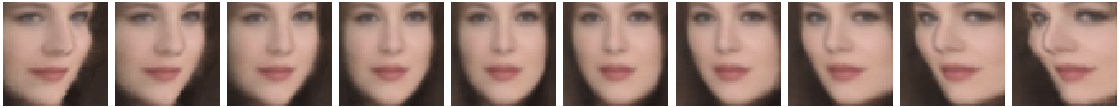}\\
 \includegraphics[width=.6\columnwidth]{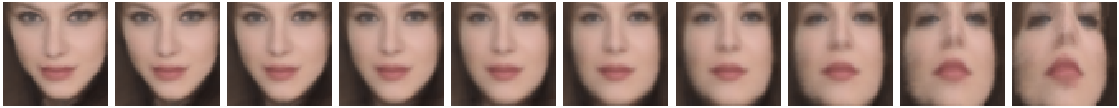}\\
 \includegraphics[width=.6\columnwidth]{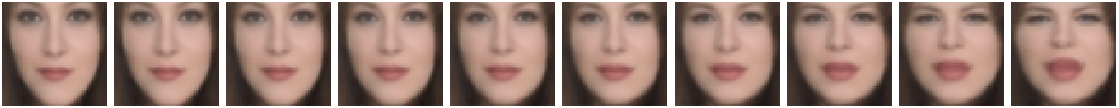}
\caption{\small Each dimension of the geometric latent vector $h_1$ encodes geometric information such as shape and viewing angle. In each row, a characteristic of the face changes from left to right. In the fist row, the shape of the face changes from wide to thin, and in the second row, the pose of the face changes from facing left to right. In the third row, the vertical tilt of the face varies from downward to upward, and in the fourth row, the face width changes from cramped to stretched. The deformable generator model is trained on the 10,000 face images randomly selected from CelebA dataset \citep{liu2015faceattributes}. The training images are cropped to $64 \times 64$ pixels, and the faces have different colors, illuminations, identities, viewing angles, shapes, and expressions.}
\label{fig:geo}
\end{center}
\end{figure}

\cite{HanLu2016}  learned the generator model by maximum likelihood.  More recently,  \cite{xing2018deformable}  generalized the model to a deformable generator model with two hidden vectors  $(h_1, h_2)$, where $h_1$ is the geometric hidden vector that generates the displacements of the pixels, or the displacement field, and $h_2$ is the appearance hidden vector that generates the appearance image before deformation. The observed image is assumed to be generated by deforming or warping the appearance image by the displacement field. Such a model can be learned by maximum likelihood, and the learned model disentangles variations in shape and appearance. 

\begin{figure}[h]
\begin{center}
 \includegraphics[width=.06\columnwidth]{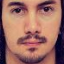}
 \hspace{1mm}
 \includegraphics[width=.06\columnwidth]{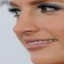}
 \includegraphics[width=.06\columnwidth]{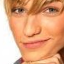}
 \includegraphics[width=.06\columnwidth]{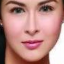}
  \includegraphics[width=.06\columnwidth]{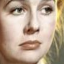}
 \includegraphics[width=.06\columnwidth]{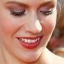}
 \includegraphics[width=.06\columnwidth]{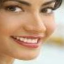}\\
 \includegraphics[width=.06\columnwidth]{./FIG/deformable/a}
 \hspace{1mm}
 \includegraphics[width=.06\columnwidth]{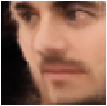}
 \includegraphics[width=.06\columnwidth]{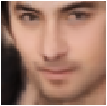}
 \includegraphics[width=.06\columnwidth]{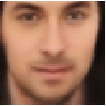}
  \includegraphics[width=.06\columnwidth]{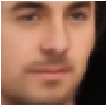}
 \includegraphics[width=.06\columnwidth]{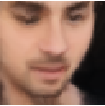}
 \includegraphics[width=.06\columnwidth]{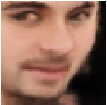}\\
 \includegraphics[width=.06\columnwidth]{./FIG/deformable/a}
 \hspace{1mm}
 \includegraphics[width=.06\columnwidth]{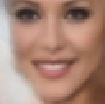}
 \includegraphics[width=.06\columnwidth]{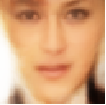}
 \includegraphics[width=.06\columnwidth]{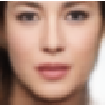}
  \includegraphics[width=.06\columnwidth]{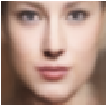}
 \includegraphics[width=.06\columnwidth]{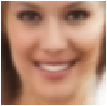}
 \includegraphics[width=.06\columnwidth]{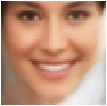}\\
\caption{\small Transferring and recombining geometric and appearance vectors. The first row shows seven faces from the CelebA data set. The second row shows the faces generated by transferring and recombining the second through seventh faces' geometric vectors $h_1$ with the first face's appearance vector $h_2$ in the first row. The third row shows the faces generated by transferring and recombining the second through seventh faces' appearance vectors $h_2$ with the first face's geometric vector $h_1$ in the first row. The deformable generator model is trained on the 10,000 face images from the CelebA data set, which are cropped to $64 \times 64$ pixels, and the faces in the training data have a wide and diverse variety of colors, illuminations, identities, viewing angles, shapes, and expressions.}
\label{fig:exchange}
\end{center}
\end{figure}

\cite{xing2018deformable}  trained the deformable generator on the 10,000 face images from the CelebA data set \citep{liu2015faceattributes}. Figure \ref{fig:geo} illustrates the change of the image if we vary the components of $h_1$, while keeping $h_2$ fixed at a certain value. Different dimensions of $h_1$ capture different aspects of shape change. Figure \ref{fig:exchange} displays an example of transferring and recombining the vectors. For two images, we can exchange their geometric vectors, so that each image changes its shape but retains its appearance. 

\begin{figure}[h]
\centering	
\hspace{0.5mm}\rotatebox{90}{\hspace{4mm}{\footnotesize obs }}	
\includegraphics[height=.08\linewidth, width=.08\linewidth]{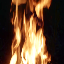}
\includegraphics[height=.08\linewidth, width=.08\linewidth]{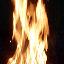}
\includegraphics[height=.08\linewidth, width=.08\linewidth]{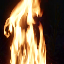}
\includegraphics[height=.08\linewidth, width=.08\linewidth]{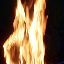}
\includegraphics[height=.08\linewidth, width=.08\linewidth]{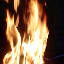}
\includegraphics[height=.08\linewidth, width=.08\linewidth]{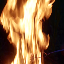}
\\
\vspace{0.5mm}
\rotatebox{90}{\hspace{4mm}{\footnotesize syn1}}
\includegraphics[height=.08\linewidth, width=.08\linewidth]{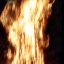}
\includegraphics[height=.08\linewidth, width=.08\linewidth]{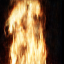}
\includegraphics[height=.08\linewidth, width=.08\linewidth]{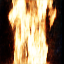}
\includegraphics[height=.08\linewidth, width=.08\linewidth]{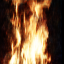}
\includegraphics[height=.08\linewidth, width=.08\linewidth]{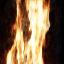}
\includegraphics[height=.08\linewidth, width=.08\linewidth]{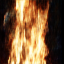}\\ 
\vspace{0.5mm}
\rotatebox{90}{\hspace{4mm}{\footnotesize syn2}}
\includegraphics[height=.08\linewidth, width=.08\linewidth]{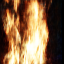}
\includegraphics[height=.08\linewidth, width=.08\linewidth]{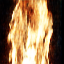}
\includegraphics[height=.08\linewidth, width=.08\linewidth]{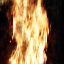}
\includegraphics[height=.08\linewidth, width=.08\linewidth]{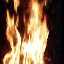}
\includegraphics[height=.08\linewidth, width=.08\linewidth]{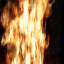}
\includegraphics[height=.08\linewidth, width=.08\linewidth]{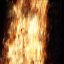}\\ 
\hspace{5mm} 
\caption{\small Generating dynamic textures. The dynamic generator model is learned by maximum likelihood from a single training video exhibiting a burning flame, sized 64 pixels $\times$ 64 pixels $\times$ 60 frames. A longer- length dynamic texture can be generated from a relatively short training sequence by just drawing longer independent and identically distributed samples from a Gaussian distribution. The first row displays 6 frames of the 60-frame observed sequence, and the second and third rows show 6 frames of two synthesized sequences of 120 frames, which are generated by the learned model. }	
\label{fig:dynamicTexture}
\end{figure}

\begin{figure}[h]
\centering	
\rotatebox{90}{\hspace{4mm}{\footnotesize obs}}	
\includegraphics[height=.08\linewidth, width=.08\linewidth]{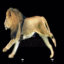}
\includegraphics[height=.08\linewidth, width=.08\linewidth]{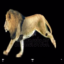}
\includegraphics[height=.08\linewidth, width=.08\linewidth]{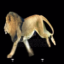}
\includegraphics[height=.08\linewidth, width=.08\linewidth]{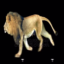}
\includegraphics[height=.08\linewidth, width=.08\linewidth]{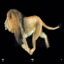}
\includegraphics[height=.08\linewidth, width=.08\linewidth]{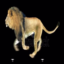}\\ \vspace{0.5mm}
\rotatebox{90}{\hspace{4mm}{\footnotesize syn1}}	
\includegraphics[height=.08\linewidth, width=.08\linewidth]{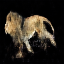} 
\includegraphics[height=.08\linewidth, width=.08\linewidth]{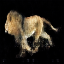} 
\includegraphics[height=.08\linewidth, width=.08\linewidth]{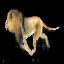}
\includegraphics[height=.08\linewidth, width=.08\linewidth]{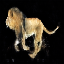} 
\includegraphics[height=.08\linewidth, width=.08\linewidth]{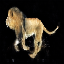}
\includegraphics[height=.08\linewidth, width=.08\linewidth]{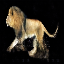} \\ \vspace{0.5mm}
\rotatebox{90}{\hspace{4mm}{\footnotesize syn2}}	
\includegraphics[height=.08\linewidth, width=.08\linewidth]{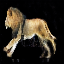} 
\includegraphics[height=.08\linewidth, width=.08\linewidth]{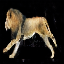}
\includegraphics[height=.08\linewidth, width=.08\linewidth]{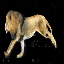} 
\includegraphics[height=.08\linewidth, width=.08\linewidth]{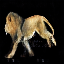} 
\includegraphics[height=.08\linewidth, width=.08\linewidth]{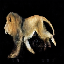}
\includegraphics[height=.08\linewidth, width=.08\linewidth]{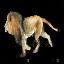}   \\
   \caption{\small Generated action patterns. The dynamic generator model is trained on an animal action data set, including 20 videos of 10 animals performing running and walking. Each observed video is scaled to 64 pixels $\times$ 64 pixels $\times$ 30 frames. The first row displays 6 frames of the observed sequence, and the second and third rows show the corresponding frames of two synthesized sequences generated by the learned model.}	
\label{fig:motion}
\end{figure}

\cite{xie2018learning} generalized the generator model to a dynamic generator model  for a video sequence $(x_t, t=1,...,T)$, where $x_t$ is an image frame at time $t$, by assuming a model of the form 
\begin{eqnarray} 
h_{t} = f_{\alpha}(h_{t-1}, z_t), \label{transition} \\ 
x_{t} = g_{\beta}(h_{t})+\epsilon_t, \label{emission} 
\end{eqnarray}
where $t=1,...,T$. Equation \ref{transition} is the transition model, and Equation \ref{emission} is the
emission model. $h_t$ is the $d$-dimensional hidden state vector, and $z_t \sim \N(0, I)$ is the noise vector of a certain dimensionality. The Gaussian noise vectors $(z_t, t = 1, ..., T)$ are independent of each other. The sequence of $(h_t, t = 1, ..., T)$ follows a nonlinear auto-regressive model, where the noise vector $z_t$ encodes the randomness in the transition from $h_{t-1}$ to $h_t$ in the $d$-dimensional state space. $f_{\alpha}$ is a feed-forward neural network or multi-layer perceptron, where $\alpha$ denotes the weight and bias parameters of the network. $x_t$ is the $D$-dimensional image, which is generated by the $d$-dimensional hidden state vector $h_t$. $g_{\beta}$ is a top-down  network, where $\beta$ denotes the weight and bias parameters of this  network. $\epsilon_t \sim \N(0, \sigma^2I_D)$ is the residual error. The model is a state-space model or hidden Markov model. \cite{xie2018learning} learned the dynmaic generator model by maximum likelihood. Figures \ref{fig:dynamicTexture} and \ref{fig:motion} show examples of learning the model from video data. Once the model is learned, we can synthesize dynamic textures from the learned model by first randomly initializing the initial hidden state $h_0$, and then following  Equations \ref{transition} and \ref{emission} to generate a sequence of images with a sequence of innovation vectors ${z_t}$ sampled from Gaussian noise distribution.

\subsection{Flow-based models} 

A flow-based model is of the form $x = g_\theta(h)$, but $h$ is of the same dimensionality as $x$, and $g_\theta$ is a composition of a sequence of simple invertible transformations, so that the probability density of $x$ can be obtained in closed form, $p_\theta(x) = p_0(g_\theta^{-1}(x)) |\partial g_\theta(x)/\partial x|^{-1}$, where $p_0$ is the density of $h$, and $|\partial g_\theta(x)/\partial x|$ is the absolute value of the determinant of the Jacobian of $g_\theta$. Such a model can be considered a special generator model with invertible mapping between the hidden vector and the signal. 

Flow-based models \citep{dinh2014nice, rezende2015variational, dinh2016density, kingma2018glow, grathwohl2018ffjord} can be traced back to independent component analysis reviewed in Section~\ref{sect:s} by, for example, \cite{dinh2014nice}. They also arise from efforts to strengthen the inference model in variational auto-encoders (VAEs) (e.g., \cite{rezende2015variational}), which are reviewed in the next section. The advantage of such models is that the normalized probability density of $x$ can be obtained in closed form, so maximum likelihood learning is simple. A disadvantage is that the mapping $g_\theta$ may be of a rather contrived form in order to ensure that the mapping is invertible and the Jacobian can be efficiently computed. 

 \section{Learning the generator model jointly with complementary models} 
 \label{sec:generator_joint}
 
 In modern deep learning literature, the generator model is usually learned jointly with a complementary model, and the learning is not based on maximum likelihood. Such learning methods are unconventional in statistics, but they can be quite powerful and can be interesting to statisticians. 
 
 \subsection{Issues with maximum likelihood} 
 
The maximum likelihood learning of the generator network in the previous section has two issues. First, the learning algorithm requires MCMC sampling of the posterior distribution $p_\theta(h|x)$ as an inner loop, which can be expensive. Second, the maximum likelihood estimator, which minimizes $\KL(p_{\rm data}\| p_\theta)$ over $\theta$, seeks to cover all the local modes of $p_{\rm data}$, and as a result,  the learned $p_\theta$ tends to be smoother than $p_{\rm data}$, and images generated by the learned $p_\theta$ tends to be less sharp than the observed images. 

To address the first issue, the VAE \citep{KingmaCoRR13,rezende2014stochastic,mnih2014neural} learns an inference model to approximate the posterior distribution. To address the second issue,  the generator model can be learned jointly with a discriminator as in generative adversarial networks (GAN) \citep{goodfellow2014generative, radford2015unsupervised} or an energy-based model that specifies the distribution of $x$ explicitly up to a normalizing constant. 

While the generator model is parameterized by a top-down network as show in the left panel of diagram (\ref{eq:diagram0}), the complementary model is parameterized by a separate bottom-up network as shown in the right panel of diagram (\ref{eq:diagram0}).  

\begin{eqnarray}
\begin{array}[c]{ccc}
  \mbox{Top-down mapping}  && \mbox{Bottom-up mapping}\\
  \mbox{hidden vector $h$} && \mbox{inference $q_\phi(h|x)$ or energy $f_\alpha(x)$}\\
 \Downarrow&&\Uparrow\\
\mbox{signal $x \approx g_\theta(h)$} && \mbox{signal $x$} \\
\mbox{(a) {Generator model} }&& \mbox{(b)  Complementary model}
\end{array}  \label{eq:diagram0}
\end{eqnarray}

 \subsection{Variational auto-encoder: joint learning with inference model}

In order to avoid MCMC sampling from the posterior $p_\theta(h|x)$, the VAE \citep{KingmaCoRR13,rezende2014stochastic,mnih2014neural} approximates $p_\theta(h|x)$ by a tractable $q_\phi(h|x)$, such as 
\begin{eqnarray}
q_\phi(h|x) \sim {\rm N}(\mu_\phi(x), {\rm diag}(v_\phi(x))), \label{eq:v}
\end{eqnarray}
where both $\mu_\phi$ and $v_\phi$ are bottom-up networks that map $x$ to $d$-dimensional vectors, with $\phi$ collecting all the weight and bias parameters of the bottom-up networks. For $h \sim q_\phi(h|x)$, we can write $h  = \mu_\phi(x) + {\rm diag}(v_\phi(x))^{1/2} z$, where $z \sim {\rm N}(0, I_d)$. Thus expectation with respect to $h \sim q_\phi(h|x)$ can be written as expectation with respect to $z$. This reparameterization trick \citep{KingmaCoRR13} helps reduce the variance in Monte Carlo integration. We may consider $q_\phi(h|x)$ as an approximation to the iterative MCMC sampling of $p_\theta(h|x)$. In other words, $q_\phi(h|x)$ is the learned inferential computation that approximately samples from $p_\theta(h|x)$. 

The VAE objective is a modification of the maximum likelihood estimation (MLE) objective: 
\begin{eqnarray}
{\rm KL}(q_{\rm data}(x) q_{\phi}(h|x) \|p_\theta(h, x))={\rm KL}(q_{\rm data}(x) \|p_\theta(x)) + {\rm KL}(q_\phi(h|x)\|p_\theta(h|x)). \label{eq:VAE}
\end{eqnarray}
We define the conditional KL divergence as $\KL(q(x|y)\|p(x|y)) = \E_{q(x, y)} [ \log (q(x|y)/p(x|y))]$ where the expectation is with respect the joint distribution $q(x, y)$. 
We estimate $\theta$ and $\phi$ jointly by 
\begin{eqnarray}
\min_\theta \min_\phi {\rm KL}(q_{\rm data}(x) q_{\phi}(h|x) \|p_\theta(h, x)), 
\end{eqnarray}
which can be accomplished by gradient descent. 
 \begin{figure}[h]
	\centering	
	\includegraphics[height=.18\linewidth]{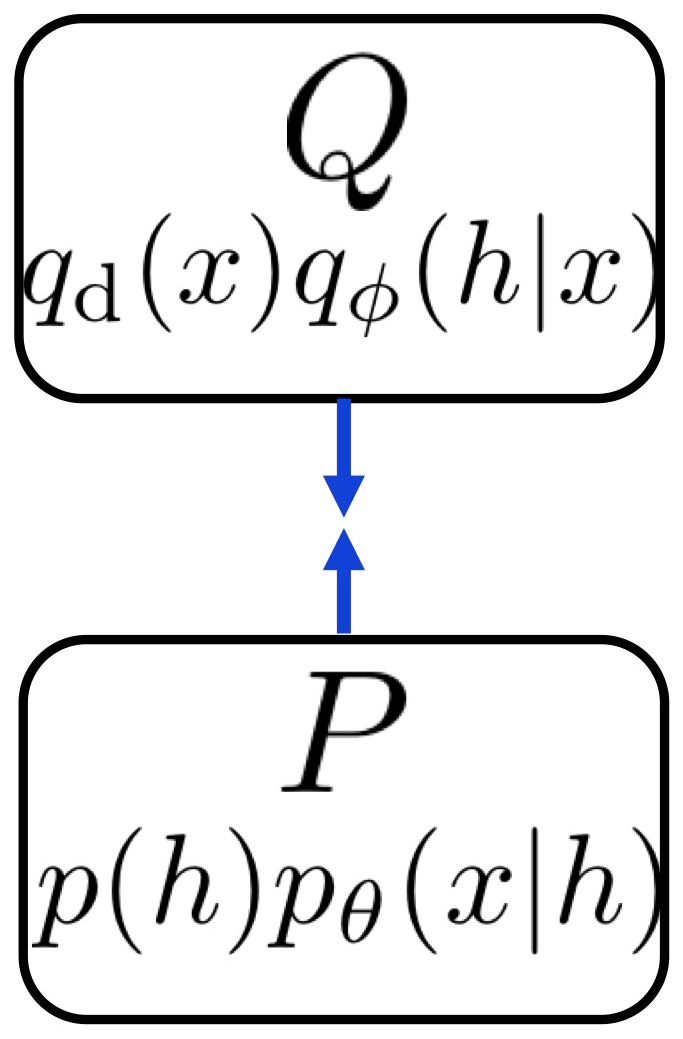} \includegraphics[height=.18\linewidth]{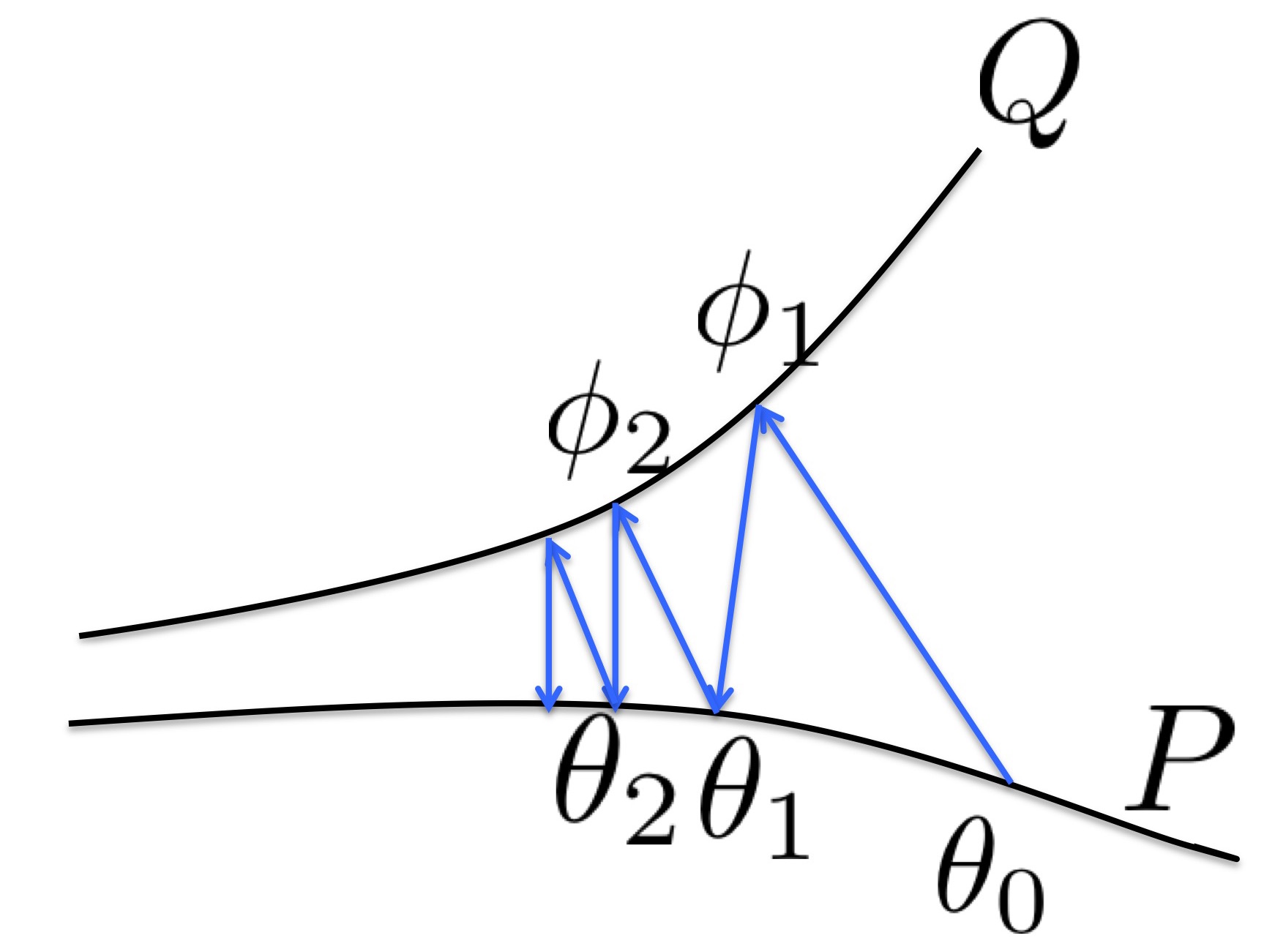} 
		\caption{\small Variational auto-encoder as joint minimization by alternating projection. $P = p(h)p_{\theta}(x|h)$ is the distribution of the complete-data model, where $p(h)$ is the prior distribution of hidden vector $h$ and $p_{\theta}(x|h)$ is the conditional distribution of $x$ given $h$. $Q = q_d(x)q_{\phi}(h|x)$ is the distribution of the complete data $(h, x)$, where $q_d(x)$ is the data distribution and $q_{\phi}(h|x)$ is the learned inferential computation that approximately samples from the posterior distribution $p_{\theta}(x|h)$. (Left) Interaction between the models. (Right) Alternating projection. The two models run toward each other.}	
	\label{fig:t2}
\end{figure}   

Define $Q(h, x) = q_{\rm data}(x) q_\phi(h|x)$. Define $P(h, x) = p(h) p_\theta(x|h)$. $Q$ is the distribution of the complete data $(h, x)$, where $q_\phi(h|x)$ can be interpreted as an imputer that imputes the missing data $h$. $P$ is the distribution of the complete-data model. The VAE is $\min_{\theta} \min_{\phi} \KL(Q\|P)$. 

 We may interpret the VAE as an alternating projection between $Q$ and $P$. (Figure \ref{fig:t2} provides an illustration. The wake-sleep algorithm \citep{hinton1995wake} is similar to the VAE, except that it updates $\phi$ by $\min_\phi {\rm KL}(P\|Q)$, where the order is flipped. 
 
  \cite{xing2018deformable}  implemented the VAE learning of the deformable generator model, and the results are similar to maximum likelihood learning. 
 
\subsubsection{Maximum likelihood estimation algorithm from the variational auto-encoder perspective} 

Recall that the MLE objective is to minimize $\KL(q_{\rm data}\|p_\theta)$. Suppose $\theta_t$ is the current estimate in the MLE algorithm. We can write 
\begin{eqnarray} 
   \KL(q_{\rm data}(x) p_{\theta_t}(h|x)\| p(h) p_\theta(x|h)) = \KL(q_{\rm data}(x) \|p_\theta(x)) + \KL(p_{\theta_t}(h|x) \| p_\theta(h|x)), 
\end{eqnarray}
 where we replace $q_\phi(h|x)$ in the VAE by $p_{\theta_t}(h|x)$. 
 
  The above identity underlies the EM algorithm \citep{dempster1977maximum}, where we find $\theta_{t+1}$ by maximizing the left-hand side over $\theta$. Because $ \KL(p_{\theta_t}(h|x) \| p_\theta(h|x))$, as a function of $\theta$, is minimized at $\theta = \theta_{t}$, with minimum value 0, $\KL(q_{\rm data}(x) p_{\theta_t}(h|x)\| p(h) p_\theta(x|h))$ majorizes $ \KL(q_{\rm data}(x) \|p_\theta(x))$ as functions of $\theta$, and both functions touch at $\theta$. Thus minimizing the left-hand side will decrease $\KL(q_{\rm data}\|p_\theta)$, which leads to the monotonicity of the EM algorithm. Moreover, the derivative of $\KL(p_{\theta_t}(h|x) \| p_\theta(h|x))$, as a function of $\theta$, is zero at $\theta_t$. Thus, the gradient of the KL divergence on the left-hand side at $\theta_t$ agrees with the gradient of the first KL divergence on the right-hand side at $\theta_t$. This leads to the identity in Equation \ref{eq:abp}. 
  
\subsubsection{Comparison with traditional variational inference} 

In the VAE, the model $q_\phi(h|x)$ and the parameter $\phi$ are shared by all the training examples $x$, so that $\mu_\phi(x)$ and $v_\phi(x)$ in Equation \ref{eq:v} can be computed directly for each $x$ given $\phi$. This is different from traditional variational inference \citep{jordan1999introduction,blei2017variational}, where for each $x$, a model $q_{\mu, v}(h)$ is learned by minimizing $\KL(q_{\mu, v}(h)\|p_\theta(h|x))$ with $x$ fixed, so that $(\mu, v)$ is computed by an iterative algorithm for each $x$, which is an inner loop of the learning algorithm. This is similar to maximum likelihood learning, except that in maximum likelihood learning, the inner loop is an iterative algorithm that samples $p_\theta(h|x)$ instead of minimizing over $(\mu, v)$. The learned networks $\mu_\phi(x)$ and $v_\phi(d)$ in the VAE are to approximate the iterative minimization algorithm by direct mappings.

\subsection{Generative adversarial net: joint learning with discriminator}
  
The generator model learned by MLE or the VAE usually cannot generate very realistic images. Both MLE and the VAE target $\KL(q_{\rm data}\|p_\theta)$, though the VAE only minimizes an upper bound of $\KL(q_{\rm data}\|p_\theta)$. Consider minimizing $\KL(q\|p)$ over $p$ within a certain model class. If $q$ is multi-modal, then $p$ is obliged to fit all the major modes of $q$ because $\KL(q\|p)$ is an expectation with respect to $q$. Thus, $p$ tends to interpolate the major modes of $q$ if $p$ cannot fit the modes of $q$ closely. As a result, $p_\theta$ learned by MLE or the VAE tends to generate images that are not as sharp as the observed images. 

The behavior of minimizing $\KL(q\|p)$ over $p$ is different from minimizing $\KL(q\|p)$ over $q$. If $p$ is multi-modal, $q$ tends to capture some major modes of $p$ while ignoring the other modes of $p$, because $\KL(q\|p)$ is an expectation with respect to $q$. In other words, $\min_q \KL(q \|p)$ encourages mode chasing, whereas $\min_p \KL(q\|p)$ encourages mode covering. 

Sharp synthesis can be achieved by GAN \citep{goodfellow2014generative, radford2015unsupervised}, which pairs a generator model $G$ with a discriminator model $D$. For an image $x$, $D(x)$ is the probability that $x$ is an observed (real) image instead of a generated (faked) image. It can be parameterized by a bottom-up network $f_\alpha(x)$, so that $D(x) = 1/(1+ \exp(-f_\alpha(x))$, i.e., logistic regression. (See Section~\ref{sect:l}). We can train the pair of $(G, D)$ by an adversarial, zero-sum game. Specifically, let $G(h) = g_\theta(h)$ be a generator. Let
 \begin{eqnarray}
      V(D, G) = \E_{\P}[\log D(X)] + \E_{h\sim p(h)}[\log (1-D(G(h))], 
 \end{eqnarray}
 where $\E_\P$ can be approximated by averaging over the observed examples, and $\E_{h}$ can be approximated by Monte Carlo average over the faked examples generated by the generator model.  We learn $D$ and $G$ by $\min_G \max_D V(D, G)$. $V(D, G)$ is the log-likelihood for $D$, i.e., the log-probability of the real and faked examples. However, $V(D, G)$ is not a very convincing objective for $G$. In practice, the training of $G$ is usually modified into maximizing $\E_{h \sim p(h)}[\log D(G(h))]$ to avoid the vanishing gradient problem.

 For a given $\theta$, let $p_\theta$ be the distribution of $g_\theta(h)$ with $h \sim p(h)$. Assuming a perfect discriminator. Then, according to the Bayes' theorem, $D(x) = \P(x)/(\P(x) + p_\theta(x))$ (assuming equal numbers of real and faked examples). Then $\theta$ minimizes the Jensen-Shannon (JS) divergence
 \begin{eqnarray}
    {\rm JS}(\P\|p_\theta) =  \KL(p_\theta\|p_{\rm mix}) + \KL(\P\|p_{\rm mix}),
  \end{eqnarray}
 where $p_{\rm mix} = (\P + p_\theta)/2$.
 
 In JS-divergence, the model $p_\theta$ also appears on the left-hand side of KL divergence. This encourages $p_\theta$ to fit some major modes of $\P$ while ignoring others. As a result, GAN learning suffers from the mode collapsing problem, i.e., the learned $p_\theta$ may miss some modes of $\P$. However, the $p_\theta$ learned by GAN tends to generate sharper images than the $p_\theta$ learned by MLE or the VAE.

\subsection{Energy-based model} 

Similar to GAN, we can pair the generator model with an energy-based model \citep{Ng2011, Dai2015ICLR, LuZhuWu2016, xieLuICML, xie2017synthesizing, xie20183dlearning, gao2018multigrid_learning}, instead of a discriminator model. Similar to the discriminator model, the energy-based model is also defined by a bottom-up network. Also similar to the discriminator model, which seeks to tell apart the images generated by the generator model and the real images, the energy-based model plays the role of an evaluator, evaluating the images generated by the generator model against the real images. We may intuitively consider the generator model as an actor or a student, and the energy-based model as a critic or a teacher. 

\subsubsection{Generalizing the exponential family model}  

The energy function in the energy-based model,  $-f_\alpha(x)$,  defines the energy of $x$, and a low energy $x$ is assigned a high probability. Specifically, we have the following probability model 
\begin{eqnarray}
     \pi_\alpha(x) = \frac{1}{Z(\alpha)} \exp\left[ f_\alpha(x) \right], 
\end{eqnarray} 
where  $f_\alpha(x)$ is parameterized by a bottom-up deep network with parameters $\alpha$, and $Z(\alpha)$ is the normalizing constant. It is the nonlinear generalization of the exponential family model (see Section \ref{sect:e}), and it is also a Gibbs distribution and a random field model. Here we drop the reference measure $\rho(x)$, or we assume it is uniform measure.  In contrast to the discriminator model $D(x)$, we may intuitively call $\pi_\alpha$ the evaluator model, where $f_\alpha$ assigns the value to $x$, and $\pi_\alpha$ evaluates $x$ by a normalized probability distribution. See the right panel of diagram (\ref{eq:diagram0}). 

In terms of learning representations, the generator  model represents the observed $x$ by a vector $h$, and the energy-based model learns multiple layers of features in the network $f_\alpha(x)$. 

The energy-based model learned by maximum likelihood tends to have stronger synthesis ability than the generator model learned by maximum likelihood, because the former directly approximates $q_{\rm data}$ by $f_\alpha$, while the latter approximates $q_{\rm data}$ by $p_\theta$ which is obtained by integrating out $h$. 

\subsubsection{Maximum likelihood} 

To learn the energy-based model $\pi_\alpha$, the maximum likelihood estimator minimizes ${\rm KL}(q_{\rm data}\|\pi_\alpha)$ over $\alpha$. We can update $\alpha$ by a gradient descent
\begin{eqnarray} 
   - \frac{\partial}{\partial \alpha} {\rm KL}(q_{\rm data}(x)\|\pi_\alpha(x)) =  \E_{q_{\rm data}} \left[\frac{\partial}{\partial \alpha} f_\alpha(x)\right] -  \E_{\pi_\alpha} \left[\frac{\partial}{\partial \alpha} f_\alpha(x)\right]. \label{eq:eb}
\end{eqnarray} 
The above identity follows from the fact that the derivative of the cumulant or log partition function $\log Z(\alpha)$  is the expectation of the derivative of $f_\alpha(x)$. 

\begin{figure}[h]
\centering
\begin{tabular}{cc}
\rotatebox{90}{\hspace{6mm}{\footnotesize obs}}
\includegraphics[width=.1\linewidth]{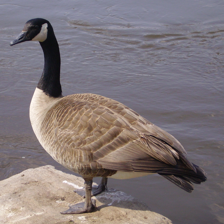} \includegraphics[width=.1\linewidth]{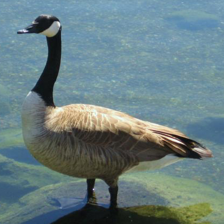}
\includegraphics[width=.1\linewidth]{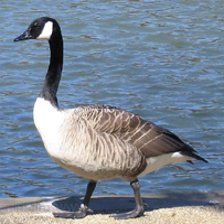}
\includegraphics[width=.1\linewidth]{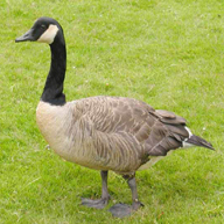}&
\rotatebox{90}{\hspace{6mm}{\footnotesize obs}}	
\includegraphics[width=.1\linewidth]{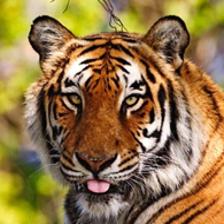}
\includegraphics[width=.1\linewidth]{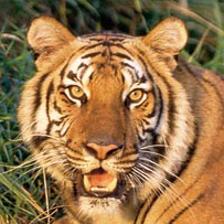}
\includegraphics[width=.1\linewidth]{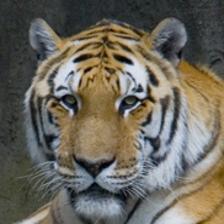}
\includegraphics[width=.1\linewidth]{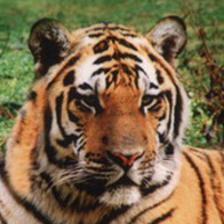}\\
\rotatebox{90}{\hspace{6mm}{\footnotesize syn}}	
\includegraphics[width=.1\linewidth]{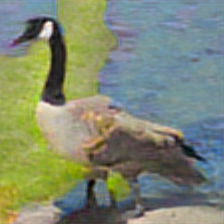}
\includegraphics[width=.1\linewidth]{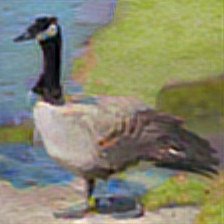}
\includegraphics[width=.1\linewidth]{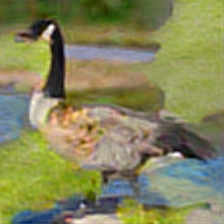}
\includegraphics[width=.1\linewidth]{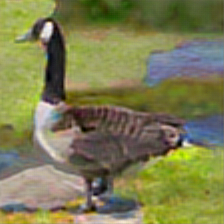}&
\rotatebox{90}{\hspace{6mm}{\footnotesize syn}}	
\includegraphics[width=.1\linewidth]{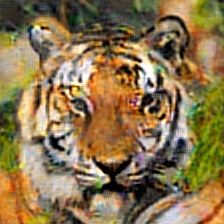}
\includegraphics[width=.1\linewidth]{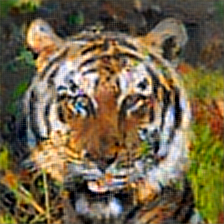}
\includegraphics[width=.1\linewidth]{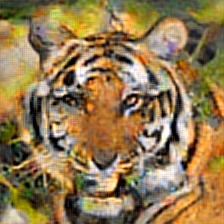}
\includegraphics[width=.1\linewidth]{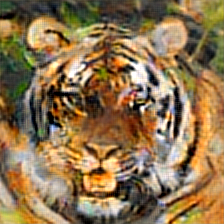}\\
(a) goose & (b) tiger
\end{tabular}
\caption{\small Learning the energy-based model by maximum likelihood: (a) goose (b) tiger. For each category, the first row displays four of the training images, and the second row displays four of the images generated by the learning algorithm. $f_{\alpha}(x)$ is parameterized by a four-layer bottom-up deep network, where the first layer has 100 $7 \times 7$ filters with sub-sampling size 2, the second layer has 64 $5 \times 5$ filters with subsampling size 1, the third layer has 20 $3 \times 3$ filters with sub-sampling size 1, and the fourth layer is a fully connected layer with a single filter that covers the whole image. The number of parallel chains for Langevin sampling is 16, and the number of Langevin iterations between every two consecutive updates of parameters is 10. The training images are $224 \times 224$ pixels. }
\label{fig:GConvNet}
\end{figure}

To implement the above update, we need to compute the expectation with respect to the current model $\pi_{\alpha}$. It can be approximated by MCMC such as Langevin dynamics or HMC that samples from $\pi_{\alpha}$. Again it can be efficiently implemented by gradient computation via back-propagation. \cite{LuZhuWu2016, xieLuICML} learned the energy-based model using such a learning method. (see Figure \ref{fig:GConvNet} for an illustration). 


More recently, \cite{Erik2019lmcmc} studied a very simple implementation of the learning algorithm where, within each learning iteration, we run $K$-step MCMC starting from a uniform noise distribution. After convergence, the $K$-step MCMC is capable of generating realistic images. 

The energy-based model is related to the discriminator model via Bayes' law (see section~\ref{sect:e} and also \cite{Dai2015ICLR, wu2018tale}). The model can be learned discriminatively by fitting a logistic regression model (see \cite{ tu2007learning, lazarow2017introspective, jin2017introspective, lee2018wasserstein}). 

\subsubsection{Adversarial contrastive divergence: joint learning of generator and energy-based model} 

To avoid MCMC sampling of $\pi_\alpha$, we may approximate it by a generator model $p_\theta$, which can generate synthesized examples directly (i.e., sampling $h$ from $p(h)$, and transforming $h$ to $x$ by $x = g_\theta(h)$). We may consider $p_\theta$ an approximation to the iterative MCMC sampling of $\pi_\alpha$. In other words, $p_\theta$ is the learned computation that approximately samples from $\pi_\alpha$, it is an approximate direct sampler of $\pi_\alpha$. 

We can learn both $\pi_\alpha$ and $p_\theta$   using the following objective function \citep{Bengio2016, dai2017calibrating}: 
\begin{eqnarray} 
\min_\alpha \max_\theta[ {\rm KL}(q_{\rm data}\|\pi_\alpha) - {\rm KL}(p_\theta \| \pi_\alpha)],  \label{eq:ACD}
\end{eqnarray} 
or equivalently 
\begin{eqnarray} 
\max_\alpha \min_\theta[ {\rm KL}(p_\theta \| \pi_\alpha) - {\rm KL}(q_{\rm data}\|\pi_\alpha)].  \label{eq:ACD1}
\end{eqnarray} 
The gradient for updating $\alpha$ becomes 
\begin{eqnarray}
\frac{\partial}{\partial \alpha} [\E_{q_{\rm data}} (f_\alpha(x)) - \E_{p_\theta}(f_\alpha(x))], \label{eq:ACD3}
\end{eqnarray}
where the intractable $\log Z(\alpha)$ term is canceled. 

Because of the negative sign in front of the second KL divergence in Equation \ref{eq:ACD}, we need $\max_\theta$ in Equation \ref{eq:ACD} or $\min_\theta$ in Equation \ref{eq:ACD1}, so that the learning becomes adversarial (illustrated in Figure \ref{fig:t3}). Inspired by {\citep{Hinton2002a},  \cite{han2018divergence}  called Equation \ref{eq:ACD} the adversarial contrastive divergence (ACD). It underlies the work of \cite{Bengio2016, dai2017calibrating}. 

     \begin{figure}[h]
	\centering	
	\includegraphics[height=.18\linewidth]{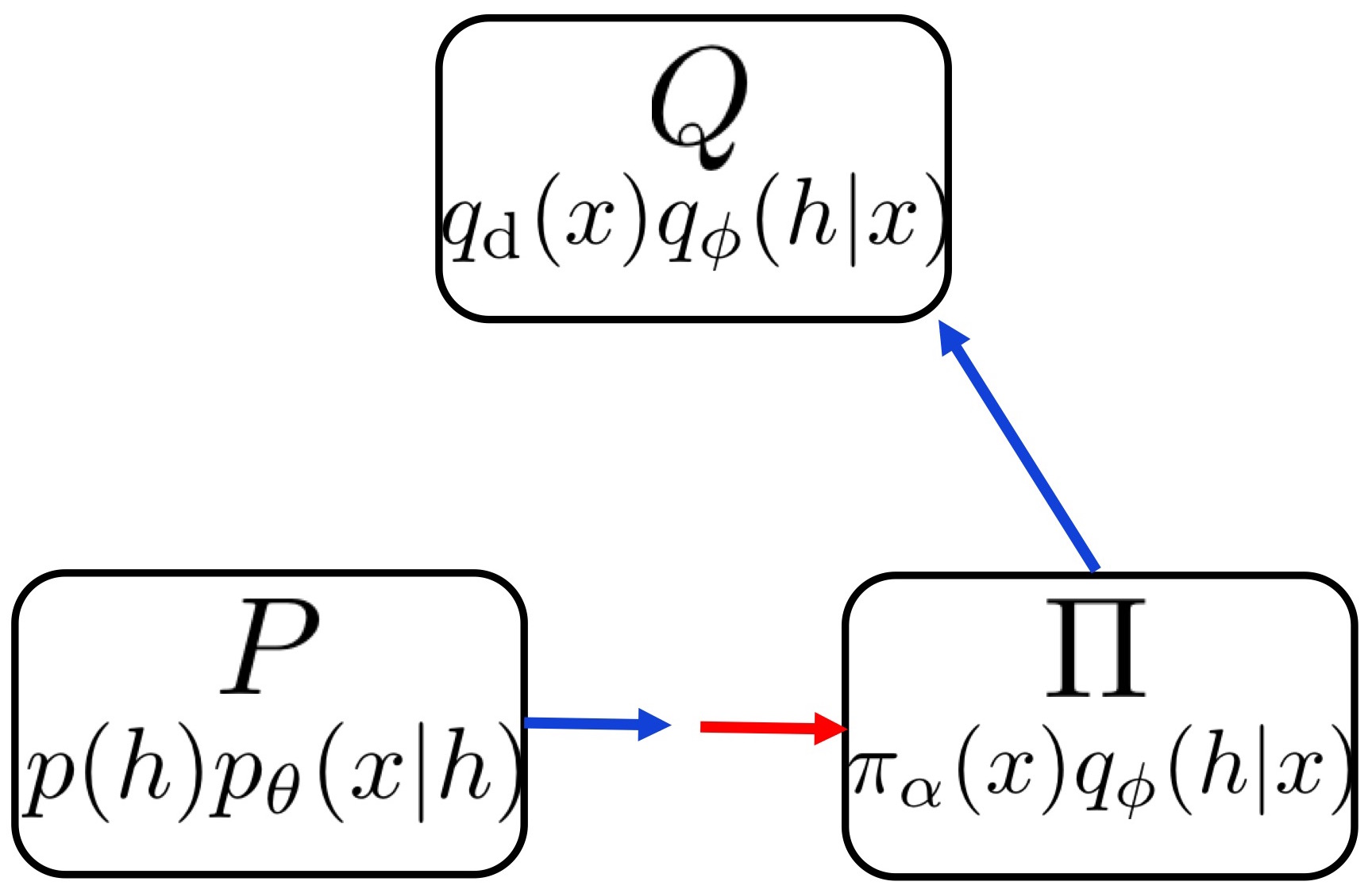} \includegraphics[height=.1\linewidth]{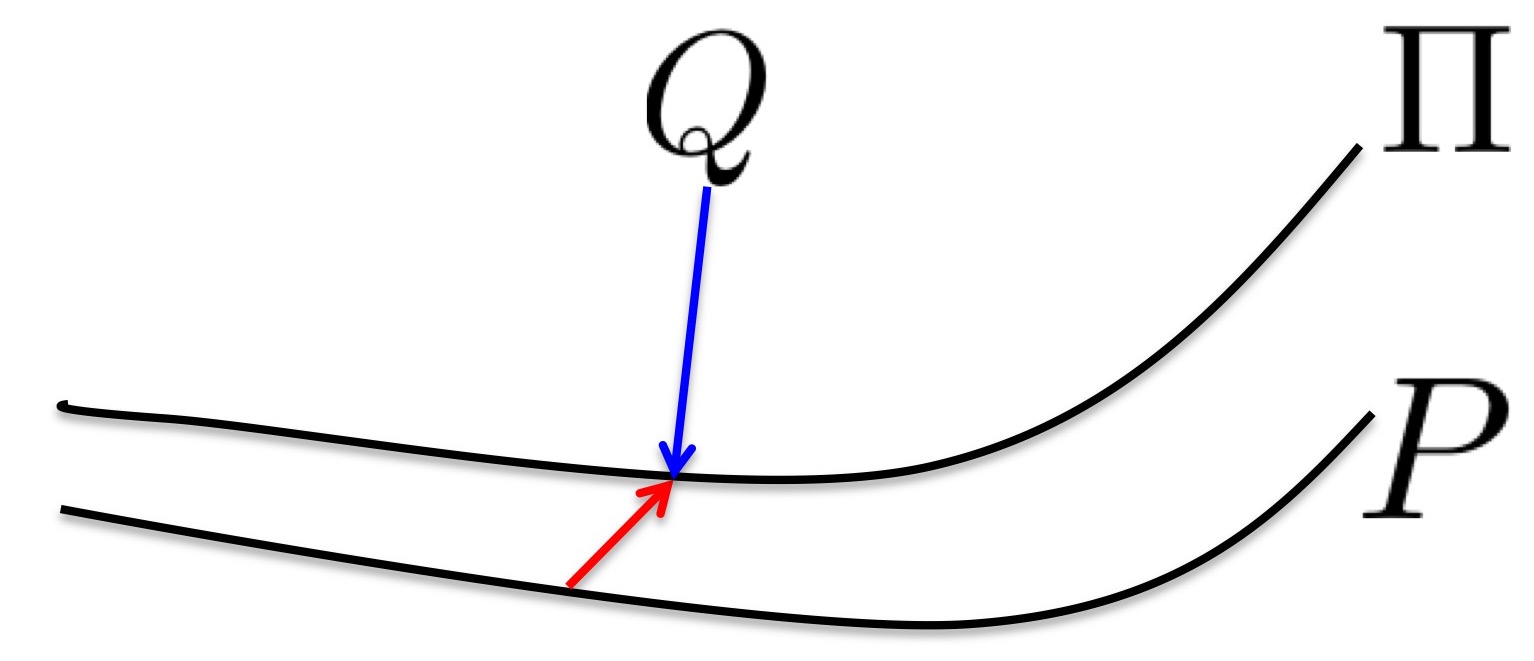} 
		\caption{\small Adversarial contrastive divergence where the energy-based model favors real data against the generator. (Left) Interaction between the models. Red arrow indicates a chasing game, where the red arrow pointing to $\Pi$ indicates
that $\Pi$ seeks to move away from $P$. The blue arrow pointing from $P$ to $\Pi$ indicates that $P$ seeks to move close to $\Pi$.
(Right) Contrastive divergence.}	
	\label{fig:t3}
\end{figure}   

The adversarial form (Equation \ref{eq:ACD} or \ref{eq:ACD1}) defines a chasing game with the following dynamics: The generator $p_\theta$ chases the energy-based model $\pi_\alpha$ in $\min_\theta {\rm KL}(p_\theta \| \pi_\alpha)$, while the energy-based model $\pi_\alpha$ seeks to get closer to $q_{\rm data}$ and away from $p_\theta$. The red arrow in Figure \ref{fig:t3} illustrates this chasing game. The result is that $\pi_\alpha$ lures $p_\theta$ toward $q_{\rm data}$. In the idealized case, $p_\theta$ always catches up with $\pi_\alpha$, and then $\pi_\alpha$ will converge to the maximum likelihood estimate $\min_\alpha \KL(q_{\rm data}\|\pi_\alpha)$, and $p_\theta$ converges to $\pi_\alpha$.  

This chasing game is different from the VAE $\min_\theta \min_\phi \KL(Q\|P)$, which defines a cooperative game where $q_\phi$ and $p_\theta$ run toward each other. 

Even though the above chasing game is adversarial, both models are running toward the data distribution. While the generator model runs after the energy-based model, the energy-based model runs toward the data distribution. As a consequence, the energy-based model guides or leads the generator model toward the data distribution. It is different from  GAN~\citep{goodfellow2014generative}, in which the discriminator eventually becomes confused because the generated data become similar to the real data. In the above chasing game, the energy-based model becomes close to the data distribution. 

The updating of $\alpha$ by Equation \ref{eq:ACD3} is similar to Wasserstein GAN (WGAN)~\citep{arjovsky2017wasserstein}, but unlike WGAN, $f_\alpha$ defines a probability distribution $\pi_\alpha$, and the learning of $\theta$  is based on $\min_\theta {\rm KL}(p_\theta \| \pi_\alpha)$, which is a variational approximation to $\pi_\alpha$. This variational approximation only requires knowing $f_\alpha(x)$, without knowing $Z(\alpha)$. However, unlike $q_\phi(h|x)$, $p_\theta(x)$ is still intractable, in particular, its entropy does not have a closed form. Thus, we can again use variational approximation, by changing the problem $\min_\theta {\rm KL}(p_\theta \| \pi_\alpha)$ to 
\begin{eqnarray}
\min_\theta \min_\phi {\rm KL}(p(h) p_\theta(x|h) \| \pi_\alpha(x) q_\phi(h|x)).
\end{eqnarray}
 Define $\Pi(h, x) =  \pi_\alpha(x) q_\phi(h|x)$, and then the problem is $\min_\theta \min_\phi {\rm KL}(P\|\Pi)$, which is analytically tractable and underlies the work of \cite{dai2017calibrating}.  In fact, 
\begin{eqnarray}
    {\rm KL}(P\|\Pi) = {\rm KL}(p_\theta(x)\|\pi_\alpha(x)) + {\rm KL}(p_\theta(h|x)\|q_\phi(h|x)). 
\end{eqnarray}
Thus, we can modify Equation \ref{eq:ACD1} into $\max_\alpha \min_\theta \min_\phi [{\rm KL}(P\|\Pi) - {\rm KL}(Q \|\Pi)]$, because  ${\rm KL}(Q\|\Pi) = \KL(q_{\rm data}\|\pi_\alpha)$. 

Note that in the VAE (Equation \ref{eq:VAE}), the objective function is in the form of KL + KL, whereas in ACD (Equation \ref{eq:ACD}), it is in the form of KL - KL. In both Equation \ref{eq:VAE} and \ref{eq:ACD}, the first KL is about maximum likelihood.  The  KL+KL form of the VAE makes the computation tractable by changing the marginal distribution of $x$ to the joint distribution of $(h, x)$. The KL-KL form of ACD makes the computation tractable by cancelling the intractable $\log Z(\alpha)$ term. Because of the negative sign in Equation \ref{eq:ACD}, the ACD objective function becomes an adversarial one or a minimax game. 

Also note that in the VAE, $p_\theta$ appears on the right-hand side of KL, whereas in ACD, $p_\theta$ appears on the left-hand side of KL. Thus in ACD, $p_\theta$ may exhibits mode chasing behavior, i.e., fitting the major modes of $\pi_\alpha$, while ignoring other modes. 

\subsubsection{Maximum likelihood estimation algorithm from the adversarial contrastive divergence perspective} 

Recall that the maximum likelihood is to minimize $\KL(q_{\rm data}\|\pi_\alpha)$. Suppose $\alpha_t$ is the current estimate of the MLE algorithm. We can consider the contrastive divergence
\begin{eqnarray} 
  \KL(q_{\rm data} \|\pi_\alpha) - \KL(\pi_{\alpha_t} \| \pi_\alpha), 
\end{eqnarray}
 where we replace $p_\theta$ in ACD by $\pi_{\alpha_t}$. Again $\KL(\pi_{\alpha_t} \| \pi_\alpha)$ as a function of $\alpha$ is minimized at $\alpha_t$, where the gradient is zero. Thus the gradient of the above contrastive divergence at $\alpha_t$ agrees with the gradient of the first KL divergence $  \KL(q_{\rm data} \|\pi_\alpha) $ for MLE, which leads to the identity in Equation \ref{eq:eb}. For the $K$-step MCMC of \cite{Erik2019lmcmc}, we can replace $\pi_{\alpha_t}$ above by the marginal distribution obtained by $K$-step MCMC toward $\pi_{\alpha_t}$, initialized at the uniform distribution. \cite{Erik2019lmcmc} also studies the learned $K$-step MCMC as a model in itself. 

\subsection{Divergence triangle: Variational auto-encoder plus adversarial contrastive divergence, joint learning of three models} 

We can combine the VAE and ACD into a divergence triangle, which involves the following three joint distributions on $(h, x)$ defined above: 
\begin{enumerate}
	\item $Q$ distribution: $Q(h, x) = q_{\rm data}(x) q_\phi(h|x)$
	\item $P$ distribution: $P(h, x) = p(h) p_\theta(x|h)$
	\item $\Pi$ distribution: $\Pi(h, x) = \pi_\alpha(x) q_\phi(h|x)$
\end{enumerate}

\begin{figure}[h]
	\centering	
	\includegraphics[width=.3\linewidth]{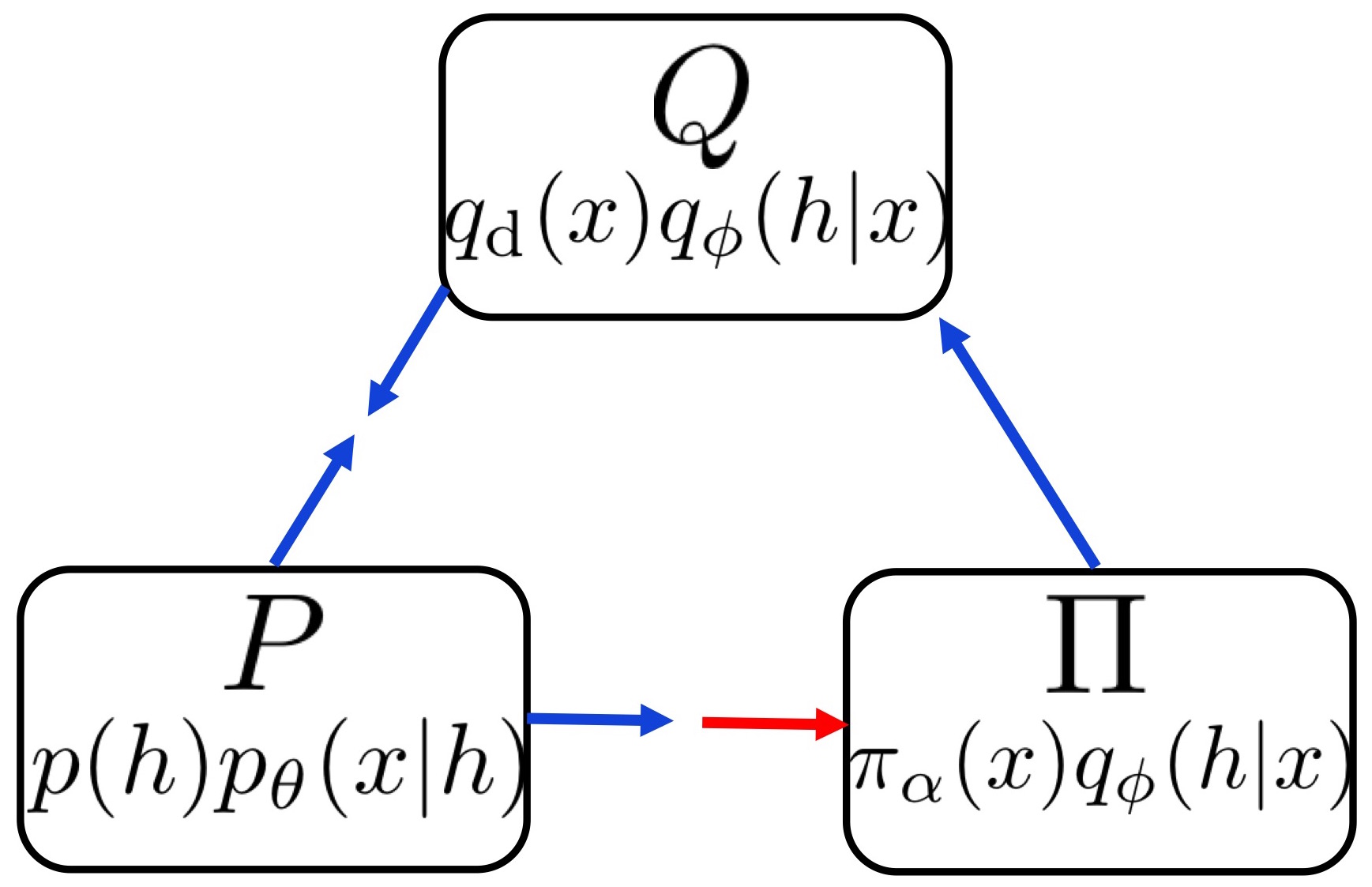} 
	\caption{\small Divergence triangle is based on the Kullback-Leibler divergences between three joint distributions, $Q$, $P$, and $\Pi$, of $(h, x)$. The blue arrow indicates the ``running toward'' behavior and the red arrow indicates the ``running away'' behavior.}	
	\label{fig:t1}
\end{figure}

 \cite{han2018divergence} proposed to learn the three models $p_\theta$, $\pi_\alpha$, and $q_\phi$ by the following divergence triangle loss functional $\D$
\begin{eqnarray} 
&& \max_\alpha \min_\theta \min_\phi \D(\alpha, \theta, \phi), \nonumber \\
&& \D = {\rm KL}(Q\|P) + {\rm KL}(P\|\Pi) - {\rm KL}(Q\|\Pi). \label{eq:triangle}
\end{eqnarray} 
See Figure \ref{fig:t1} for an illustration. The divergence triangle is based on the three KL divergences between the three joint distributions on $(h, x)$. It has a symmetric and anti-symmetric form, where the anti-symmetry is due to the negative sign in front of the last KL divergence and the maximization over $\alpha$. Compared to the VAE and ACD objective functions in the previous subsections, ${\rm KL}(Q\|P)$ is the VAE part, and $ {\rm KL}(P\|\Pi) - {\rm KL}(Q\|\Pi)$ is the ACD part. 

The divergence triangle leads to the following dynamics between the three models: (a) $Q$ and $P$ seek to get close to each other. (b) $P$ seeks to get close to $\Pi$.  (c) $\pi$ seeks to get close to $q_{\rm data}$, but it seeks to get away from $P$, as indicated by the red arrow. Note that  $\KL(Q\|\Pi) = \KL(q_{\rm data}\|\pi_\alpha)$, because $q_\phi(h|x)$ is canceled out. The effect of (b) and (c) is that $\pi$ gets close to $q_{\rm data}$ while inducing $P$ to get close to $q_{\rm data}$ as well, or in other words, $P$ chases $\pi_\alpha$ toward $q_{\rm data}$. 

 \cite{han2018divergence}  also employed the layer-wise training scheme of \cite{karras2017progressive} to learn models by divergence triangle from the CelebA-HQ data set \citep{karras2017progressive}, including 30,000 celebrity face images with resolutions of up to $1,024 \times 1,024$ pixels. The learning algorithm converges stably, without extra tricks, to obtain realistic results as shown in Figure \ref{fig:a1}. 
 
 Figure \ref{fig:a1}a  displays a few $1,024 \times 1,024$ images generated by the learned generator model with 512-dimensional latent vector. Figure \ref{fig:a1}b shows an example of interpolation. The two images at the two ends are generated by two different latent vectors. The images in between are generated by the vectors that are linear interpolations of the two vectors at the two ends. Even though the interpolation is linear in the latent vector space, the nonlinear mapping leads to a highly nonlinear interpolation in the image space. We first do linear interpolation between the latent vectors at the two ends, i.e.,$(1-\alpha) \times h_0+ \alpha \times h_1$, where $h_0$ and $h_1$ are two latent vectors at two ends, respectively, and $\alpha$ is in the closed unit interval [0, 1]. The images in between are generated by mapping those interpolated vectors to image space via the learned generator.  The interpolation experiment shows that the algorithm can learn a smooth generator model that traces the manifold of the data distribution.

\begin{figure}[h]
	\centering	
	(a) Generated face images\\
	\includegraphics[width=.99\linewidth]{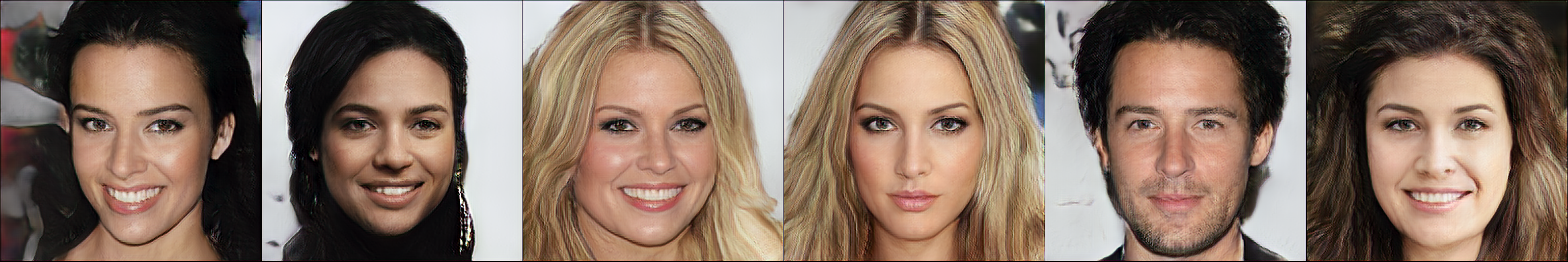} \\
	(b) Linear interpolations\\
	\includegraphics[width=.99\linewidth]{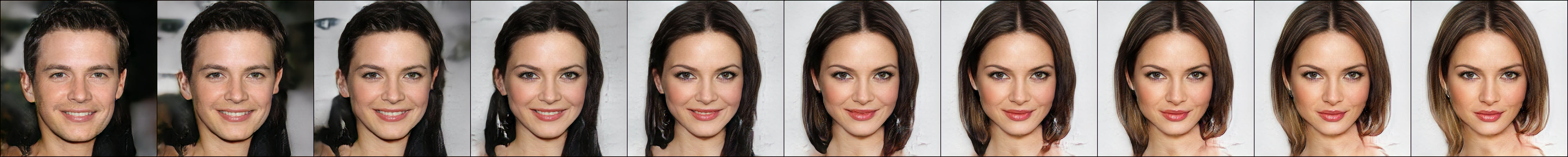}
		\caption{\small Learning generator model by divergence triangle from the CelebA-HQ data set \citep{karras2017progressive} that includes 30,000 high-resolution celebrity face images. (a) Generated face images with $1,024 \times 1,024$ resolution  sampled from the learned generator model with 512-dimensional latent vector. (b) Linear interpolation of the vector representations. The images at the two ends are generated from latent vectors randomly sampled from a Gaussian distribution. Each image in the middle is obtained by first interpolating the
two vectors of the two end images, and then generating the image using the generator.}	
	\label{fig:a1}
\end{figure}

\subsection{Cooperative learning via MCMC teaching} \label{sect:coop}

In ACD, the generator model $p_{\theta}$ is used to approximate the energy-based model $\pi_{\alpha}$, and we treat the examples generated by $p_{\theta}$ as if they are generated from $\pi_{\alpha}$ for the sake of updating $\alpha$. The gap between $p_{\theta}$ and $\pi_{\alpha}$ can cause bias in learning.  In the work of \cite{CoopNets2016, coopnets2018}, we proposed to bring back MCMC to bridge the gap. Instead of running MCMC from scratch, we run a finite-step MCMC toward $\pi_{\alpha}$, initialized from the examples generated by $p_{\theta}$. We then use the examples produced by the finite-step MCMC as the synthesized examples from $\pi_{\alpha}$ for updating $\alpha$. Meanwhile we update $p_{\theta}$ based on how the finite-step MCMC revises the initial examples generated by $p_{\theta}$; in other words, the energy-based model (as a teacher) $\pi_{\alpha}$ distills the MCMC into the generator (as a student) $p_{\theta}$. We call this scheme  cooperative learning. 

Specifically, we first generate $\hat{h}_i \sim \N(0, I_d)$, and then generate $\hat{x}_i = g_{\theta}(\hat{h}_i) + \epsilon_i$, for $i = 1,..., \tilde{n}$. Starting from $\{\hat{x}_i, i = 1,...,\tilde{n}\}$, we run MCMC such as Langevin dynamics for a finite number of steps toward $\pi_{\alpha}$ to get $\{\tilde{x}_i, i = 1,...,\tilde{n}\}$, which are revised versions of $\{\hat{x}_i\}$.  $\{\tilde{x}_i\}$ are used as the synthesized examples from the energy-based model. We can then update $\alpha$ according to Equation \ref{eq:eb}. 

The energy-based model can teach the generator via MCMC. The key is that in the generated examples, the latent $h$ is known. In order to update $\theta$ of the generator model, we treat $\{\tilde{x}_i, i = 1,...,\tilde{n}\}$ as the training data for the generator. Since these $\{\tilde{x}_i\}$ are obtained by the Langevin dynamics initialized from $\{\hat{x}_i\}$, which are generated by the generator model with known latent factors $\{\hat{h}_i\}$, we can update $\theta$ by learning from the complete data $\{(\hat{h}_i, \tilde{x}_i); i = 1,...,\tilde{n}\}$, which is a supervised learning problem, or more specifically, a nonlinear regression of $\tilde{x}_i$ on $\hat{h}_i$. At $\theta^{(t)}$, the latent factors $\hat{h}_i$ generates and thus reconstructs the initial example $\hat{x}_i$. After updating $\theta$, we want $\hat{h}_i$ to reconstruct the revised example $\tilde{x}_i$. That is, we revise $\theta$ to absorb the MCMC transition from $\hat{x}_i$ to $\tilde{x}_i$. The left panel of diagram (\ref{eq:diagram_coop}) illustrates the basic idea.

\begin{eqnarray}
\begin{tikzpicture}
  \matrix (m) [matrix of math nodes,row sep=3em,column sep=4em,minimum width=2em]
  {
     \hat{h}_i & \\
      \hat{x}_i & \tilde{x}_i \\};
  \path[-stealth]
    (m-1-1) edge [double] node [left] {$\theta^{(t)}$} (m-2-1)
        (m-1-1) edge [double] node [right] {$\;\;\theta^{(t+1)}$} (m-2-2)      
            (m-2-1) edge [dashed]  node [below] {$\alpha^{(t)}$} (m-2-2);
                     \end{tikzpicture}  
            \begin{tikzpicture}
  \matrix (m) [matrix of math nodes,row sep=3em,column sep=4em,minimum width=2em]
  {
     \hat{h}_i & \tilde{h}_i \\
      \hat{x}_i & \tilde{x}_i \\};
  \path[-stealth]
      (m-1-1) edge  [dashed]  node [above] {$\theta^{(t)}$} (m-1-2)
    (m-1-1) edge [double] node [left] {$\theta^{(t)}$} (m-2-1)
        (m-1-2) edge [double]  node [right] {$\theta^{(t+1)}$} (m-2-2)
          (m-2-1) edge  [dashed]   node [below] {$\alpha^{(t)}$} (m-2-2);
            \end{tikzpicture}      
            \label{eq:diagram_coop} 
            \end{eqnarray}

In the two diagrams in (\ref{eq:diagram_coop}), the double-line arrows indicate generation and reconstruction by the generator model, while the dashed-line arrows indicate Langevin dynamics for MCMC sampling and inference in the two models. The right panel of diagram (\ref{eq:diagram_coop}) illustrates a more rigorous method, where we initialize the MCMC for inferring $\{\tilde{h}_i\}$ from the known $\{\hat{h}_i\}$ and then update $\theta$ based on $\{(\tilde{h}_i, \tilde{x}_i), i = 1,...,\tilde{n}\}$.


The theoretical understanding of the cooperative learning scheme is given below.

(1) Modified contrastive divergence for the energy-based model. In the traditional contrastive divergence \citep{Hinton2002a}, $\hat{x}_i$ is taken to be the observed $x_i$. In cooperative learning, $\hat{x}_i$ is generated by $p_{\theta^{(t)}}$. Let $M_\alpha$ be the Markov transition kernel of finite steps of Langevin dynamics that samples $\pi_\alpha$. Let $(M_{\alpha}p_{\theta})(x) =\int M_{\alpha}(x{'}, x)p_{\theta}(x{'})dx{'}$ be the marginal distribution by running $M_{\alpha}$ initialized from $p_{\theta}$. Then similar to the traditional contrastive divergence, the learning gradient of the evaluator model $\alpha$ at iteration $t$ is the gradient of $\KL(\P \parallel \pi_\alpha) - \KL(M_{\alpha^{(t)}} p_{\theta^{(t)}} \parallel \pi_\alpha)$ with respect to $\alpha$. In the traditional contrastive divergence, $\P$ takes the place of $p_{\theta^{(t)}}$ in the second KL divergence.

(2) MCMC teaching of the generator model. The learning gradient of the generator $\theta$ in the right panel of diagram (\ref{eq:diagram_coop}) is the gradient of  $\KL(M_{\alpha^{(t)}} p_{\theta^{(t)}} \parallel p_\theta)$ with respect to $\theta$. Here $\pi^{(t+1)} = M_{\alpha^{(t)}} p_{\theta^{(t)}}$ takes the place of $\P$ as the data to train the generator model. It is much easier to minimize  $\KL(M_{\alpha^{(t)}} p_{\theta^{(t)}}\parallel p_\theta)$ than to minimize $\KL(\P \parallel p_\theta)$ because the latent variables are essentially known in the former, so the learning is supervised. The MCMC teaching alternates between Markov transition from $p_{\theta^{(t)}}$ to $\pi^{(t+1)}$,  and projection from $\pi^{(t+1)}$ to $p_{\theta^{(t+1)}}$, as illustrated by Figure \ref{fig:LP}. 

\begin{figure}[h]
\begin{center}
\includegraphics[height=.18\linewidth]{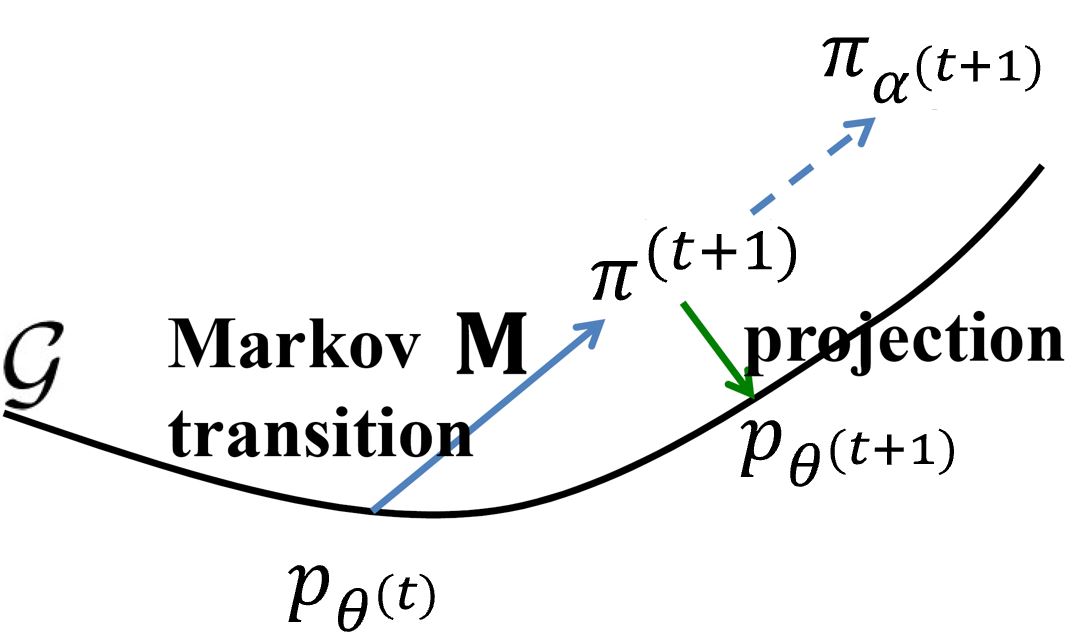}
\caption{\small The MCMC teaching of the generator alternates between Markov transition and projection. The family of the generator models $\G$ is illustrated by the black curve, and each distribution is illustrated by a point. $p_{\theta}$ is a generator model, and $\pi_{\alpha}$ is an energy-based model.}
\label{fig:LP}
\end{center}
\end{figure}

Figure \ref{fig:CoopNets} displays two examples of image synthesis by cooperative learning algorithm on datasets, LSUN bedrooms \citep{yu2015lsun} and CelebA human faces \citep{liu2015faceattributes}.

\begin{figure}[h]
\centering
\begin{tabular}{cc}
\rotatebox{90}{\hspace{10mm}{\footnotesize obs}}
\includegraphics[width=0.45\linewidth]{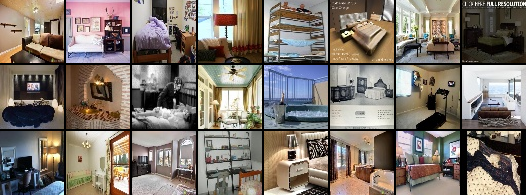} &
\rotatebox{90}{\hspace{10mm}{\footnotesize obs}}
\includegraphics[width=0.45\linewidth]{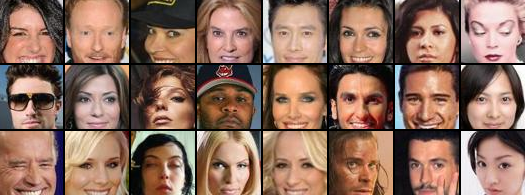}\\
\rotatebox{90}{\hspace{10mm}{\footnotesize syn}}	
\includegraphics[width=0.45\linewidth]{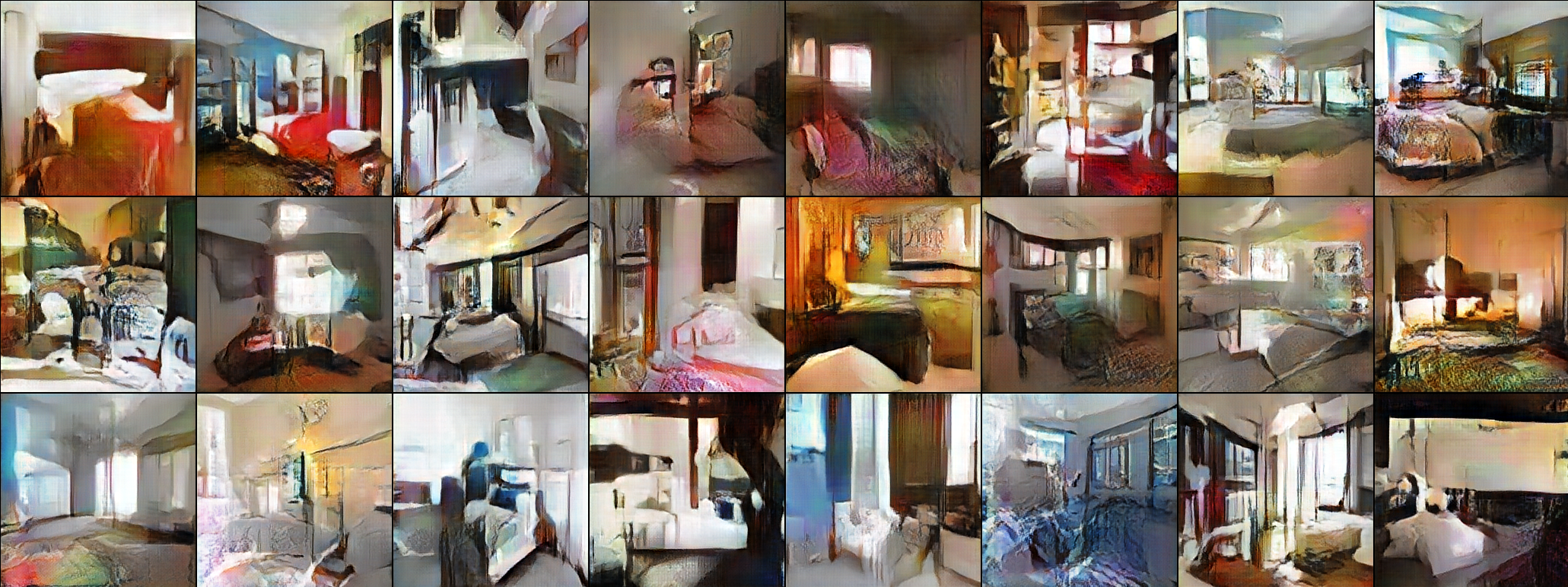} &
\rotatebox{90}{\hspace{10mm}{\footnotesize syn}}	
\includegraphics[width=0.45\linewidth]{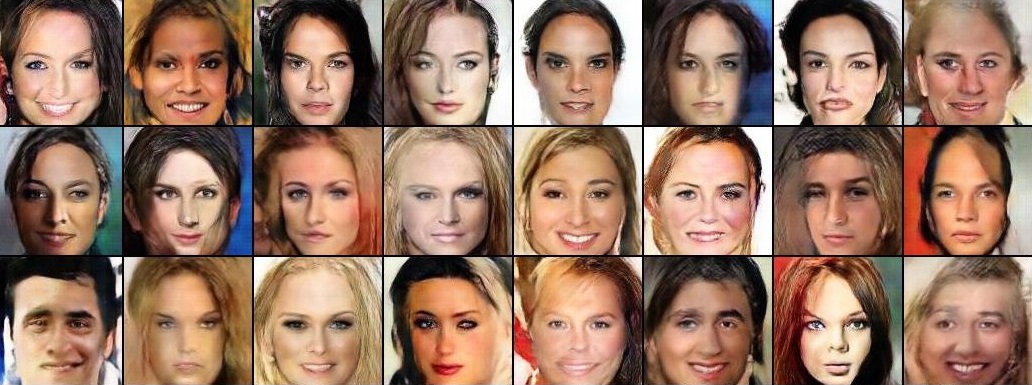}\\
(a) bedroom & (b) face
\end{tabular}
\caption{\small Image synthesis by cooperative learning. (a) Generating bedroom images ($256 \times 256$ pixels). The synthesized images are generated by the cooperative learning algorithm that learns from the LSUN data set \citep{yu2015lsun} with 3,033,000 training images. (b) Generating human face images ($128 \times 128$ pixels). The synthesized images are generated by the cooperative learning algorithm that learns from the CelebA data set \citep{liu2015faceattributes} with 200,000 training images. For each category, the top panel shows examples of the training images, and the bottom panel shows examples of the synthesized images generated by the learned models.}
\label{fig:CoopNets}
\end{figure}

\section{Learning conditional generator model} 
\label{sec:conditional}
The models and methods in the previous section can be easily generalized to conditional versions, which can be more useful in various applications. 

\subsection{Conditional generators, conditional variational auto-encoders and conditional generative adversarial networks} 

The unconditioned generator model can be extended to a conditional model. Let $x$ be the observed signal, and $c$ be the observed condition. For instance, $x$ may be an image, and $c$ may be a class label (e.g., cat or bird), or some text description (e.g., a bird is flying). The goal is to learn the conditional distribution $p_{\theta}(x|c)$ of the signal $x$ given the condition $c$ from the training data set of the pairs $\{(x_i, c_i), i = 1, ..., n\}$ that follow the data distribution $q_{\rm data}(x, c)$. This is a supervised learning problem, except that $x$ is a high-dimensional signal, and $c$ may also be high dimensional. 

The conditional generator model is of the following form:
\begin{eqnarray} 
h \sim {\rm N}(0, I_d), \; x = g_\theta(h, c) + \epsilon,
\end{eqnarray} 
where $g_\theta(h,c)$ is a top-down convolutional network (ConvNet) defined by the parameters $\theta$. The ConvNet $g$ maps the latent noise vector $h$ together with the observed condition $c$ to the signal $x$ directly. Again, $\epsilon \sim {\rm N}(0, \sigma^2 I_D)$ is the residual noise signal that is independent of $h$. If $c$ is the class label, it takes the form as a one-hot vector of label and is concatenated with $h$ and fed into the decoder $g$. If the $c$ is of high dimensionality, e.g., an image or text, we can parameterize $g$ by an encoder-decoder structure: We first encode $c$ into a latent vector $z$, and then we map the concatenation of $h$ and $z$, i.e., $(h, z)$, to $x$ by a decoder. Given $c$, we can generate $x$ from the conditional generator model by direct sampling, i.e., first sampling $h$ from its prior distribution, and then mapping $(h,c)$ into $x$ directly.

The conditional generator model can be trained by maximum likelihood or, equivalently, minimizing the KL divergence ${\rm KL}(q_{\rm data}(x|c)\| p_\theta(x|c))$ over $\theta$.  The gradient of the conditional log-likelihood is computed by 
\begin{eqnarray} 
    - \frac{\partial}{\partial \theta} {\rm KL}(q_{\rm data}(x|c)\|p_\theta(x|c)) = \E_{q_{\rm data}(x, c) p_\theta(h|x,c)} \left[\frac{\partial}{\partial \theta} \log p_\theta(h, x|c)\right], 
    \label{eq:abp_cond}
\end{eqnarray} 
where the expectation with respect to the conditional posterior distribution $p_\theta(h|x,c)$ can be approximated via MCMC sampling of $p_\theta(h|x,c)$.

Conditional VAEs \citep{sohn2015learning} train the conditional generator model by learning a tractable conditional inference model $q_\phi(h|x,c)$ to approximate the true conditional posterior distribution $p_\theta(h|x,c)$ for the sake of getting around the MCMC sampling from $p_\theta(h|x,c)$. Its objective function is given by 
\begin{eqnarray}
{\rm KL}(q_{\rm data}(x|c) q_{\phi}(h|x,c) \|p_\theta(h, x|c))={\rm KL}(q_{\rm data}(x|c) \|p_\theta(x|c)) + {\rm KL}(q_\phi(h|x,c)\|p_\theta(h|x,c)). \label{eq:cVAE}
\end{eqnarray} 

The adversarial learning framework can also be  used to train the conditional generator model, where both the generator and discriminator are conditioned on the same condition. The resulting model is called conditional GAN \citep{mirza2014conditional}, whose objective function of a two-player minimax game is
 \begin{eqnarray}
      V(D, G) = \E_{\P}[\log D(x|c)] + \E_{h\sim p(h)}[\log (1-D(G(h|c))]. 
      \label{eq:gan_cond}
 \end{eqnarray}

The conditional generator models have a wide variety of application scenarios in computer vision and graphics, such as synthesizing images from text description \citep{reed2016generative}, image-to-image translation \citep{isola2017image} including synthesizing photo images from label maps or edge maps, and video-to-video translation \citep{wang2018video} including converting an input source video, e.g., a sequence of semantic segmentation masks, to a target realistic video.

\subsection{Conditional learning via fast thinking initializer and slow thinking solver}

Recently, \cite{xie2019multimodal} extended the cooperative learning scheme to the conditional learning problem by jointly learning a conditional energy-based model and a conditional generator model. The conditional energy-based model is of the following form
\begin{eqnarray}
 \pi_{\alpha}(x|c)= \frac{1}{Z(c, \alpha)}\exp[f_{\alpha}(x,c)],
\end{eqnarray}
where $x$ is the input signal and $c$ is the condition. $Z(c, \alpha)$ is the normalizing constant conditioned on $c$. $f_{\alpha}(x, c)$ can be defined by a bottom-up ConvNet where $\alpha$ collects all the weight and bias parameters. Fixing the condition $c$, $f_{\alpha}(x, c)$ defines the value of $x$ for the condition $c$, and $-f_{\alpha}(x, c)$ defines the conditional energy function. 
$\pi_\alpha(x|c)$ is also a deep generalization of conditional random fields \citep{lafferty2001conditional}. 
Both the conditional generator model and the conditional energy-based model can be learned jointly by the cooperative learning scheme in Section~\ref{sect:coop}.

Figure \ref{fig:mnist_cond} shows some examples of learning the conditional distribution of an image given a class label. The two models are jointly learned on 30,000 handwritten digit images from the MNIST database \citep{lecun1998gradient} conditioned on their class labels, which are encoded as one-hot vectors. For each class, 10 randomly sampled images are displayed. Each column is conditioned on one label and each row is a different generated sample.

\begin{figure}[h]
\centering	
\includegraphics[width=.3\linewidth]{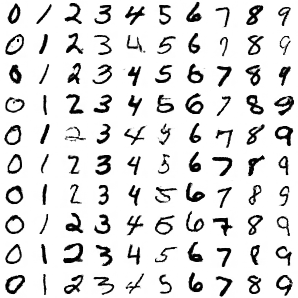}  
\caption{\small Generated handwritten digits conditioned on class labels. Each column is conditioned on one class label, and each row represents a different generated handwritten digit image. The synthesized images are generated by the jointly trained initializer and solver from 30,000 handwritten digit images along with their class labels from the MNIST database. The image size is $64 \times 64$ pixels. Abbreviation: MNIST, Modified National Institute of Standards and Technology.}%
\label{fig:mnist_cond}
\end{figure}


Figure \ref{fig:recovery} shows some examples of pattern completion on the CMP (Center for Machine Perception) Facades data set \citep{tylevcek2013spatial} by learning a mapping from an occluded image ($256 \times 256$ pixels), where a mask of the size of $128 \times 128$ pixels is centrally placed onto the original version, to the original image. In this case, $c$ is the observed part of the signal, and $x$ is the unobserved part of the signal.

\begin{figure}[h]
\centering
\begin{tabular}{ccccc}
{\small input} & {\small ground truth} & {\small initializer} & {\small solver} & {\small conditional GAN}\\ 
\includegraphics[width=.13\linewidth]{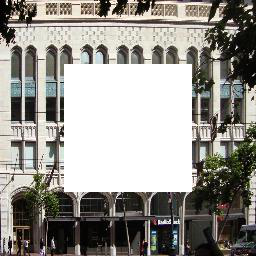}& 
\includegraphics[width=.13\linewidth]{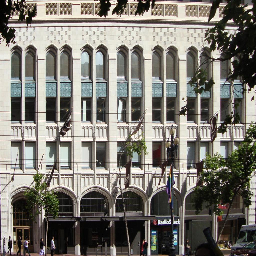}&
\includegraphics[width=.13\linewidth]{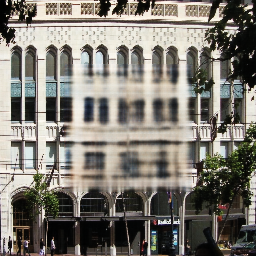}&
\includegraphics[width=.13\linewidth]{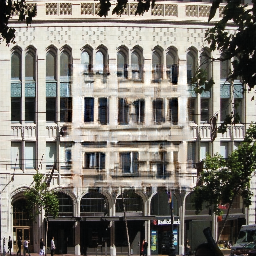}&
\includegraphics[width=.13\linewidth]{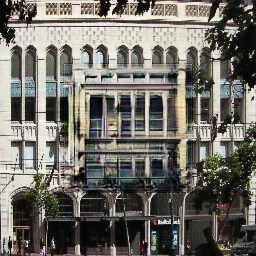}\\ \vspace{1mm}
\includegraphics[width=.13\linewidth]{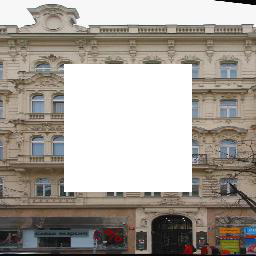}& 
\includegraphics[width=.13\linewidth]{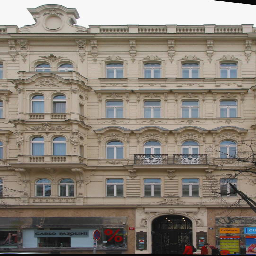}&
\includegraphics[width=.13\linewidth]{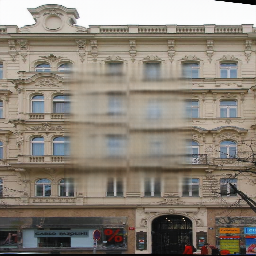}&
\includegraphics[width=.13\linewidth]{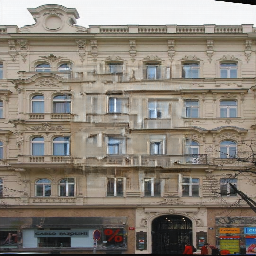}&
\includegraphics[width=.13\linewidth]{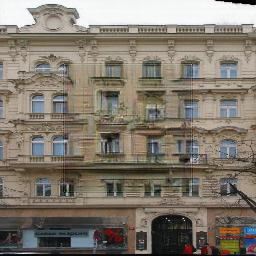}\\ \vspace{1mm}
\includegraphics[width=.13\linewidth]{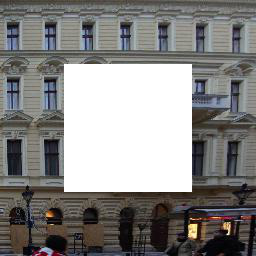}& 
\includegraphics[width=.13\linewidth]{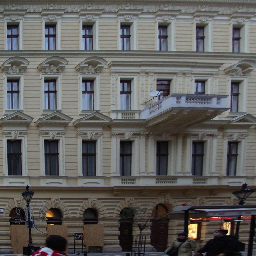}&
\includegraphics[width=.13\linewidth]{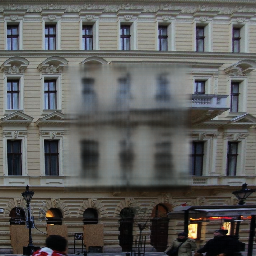}&
\includegraphics[width=.13\linewidth]{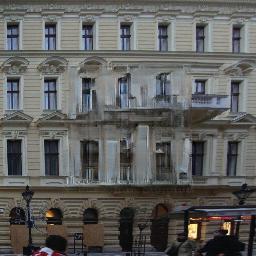}&
\includegraphics[width=.13\linewidth]{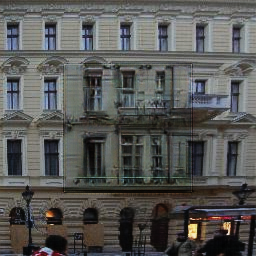}\\
\end{tabular}
\caption{\small Pattern completion by conditional learning. Each row displays one example. The first image is the testing image (256 $\times$ 256 pixels) with a hole of $128 \times 128$ that needs to be recovered, the second image shows the ground truth, the third image shows the result recovered by the initializer (i.e., conditional generator model), the fourth image shows the result recovered by the solver (i.e., the MCMC sampler of the conditional energy-based model, initialized from the result of the initializer), and the last image shows the result recovered by the conditional GAN as a comparison.}
\label{fig:recovery}
\end{figure}

The cooperative learning of the conditional generator model and conditional energy-based model can be interpreted as follows. The conditional energy function defines the objective function or value function, i.e., it defines what solutions are desirable given the condition or the problem. The solutions can then be obtained by an iterative optimization or sampling algorithm such as MCMC. In other words, the  conditional energy-based model leads to a solver in the form of an iterative algorithm, and this iterative algorithm is a slow thinking process. In contrast, the conditional generator model defines a direct mapping from condition or problem to solutions, and it is a fast thinking process. We can use the fast thinking generator as an initializer to generate the initial solution, and then use the slow thinking solver to refine the fast thinking initialization by the iterative algorithm. The cooperative learning scheme enables us to learn both the fast thinking initializer and slow thinking solver. Unlike conditional GAN, the cooperative learning scheme has a slow thinking refining process, which can be important if the fast thinking initializer is not optimal. 

In terms of inverse reinforcement learning \citep{abbeel2004apprenticeship, ziebart2008maximum}, the conditional energy-based model defines the reward or value function, and the iterative solver defines an optimal control or planning algorithm. The conditional generator model defines a policy. The fast thinking policy is about habitual, reflexive, or impulsive behaviors, while the slow thinking solver is about deliberation and planning. Compared with the policy, the value is usually simpler and more generalizable, because it is in general easier to specify what one wants than to specify how to produce what one wants. 

\section{Conclusions} 

This article reviews recent work on learning representations from a statistical perspective. We focus on unsupervised learning from unlabeled data. The representations can be either generative, like factor analysis, or relative, like multidimensional scaling. 

A generative representation is a latent variable model. In this article, we focus on learning the model with a hidden vector at the top layer, and the hidden vector generates the signal via a linear or nonlinear transformation. Such a model can and should be extended to multiple layers of hidden vectors, or a hierarchical or graphical model \citep{lee2009convolutional,salakhutdinov2009deep}. While statisticians tend to learn such models by maximum likelihood or Bayesian methods, with the help of MCMC, researchers in deep learning prefer to learn such models by variational approximations or adversarial training. It is our hope that this article explains the latter methods and connects them to more traditional statistical methods. 

A relative representation seeks to preserve important relations in the original observations. Such representations can be useful for exploratory data analysis or visualization. In relative representations, matrix representations can be used to represent the relations. For modeling dynamic systems, we can use vectors to represent the states, and matrices to represent the changes of states caused by motions and actions. 

Comparing vector representations and matrix representations, the latter are much less studied than the former, but the brain appears to need both for representing the sensory data---vector representations are ``nouns'' and matrix representations are ``verbs''. From a philosophical point of view, the brain only has access to the sensory data (including external and internal sensory data), and our notions of the outside world are the vector and matrix representations that the brain invents to explain the sensory data. In a sense, only data are real, and the outside world as we see it is more imaginary than real. 

{\small

\subsubsection*{Acknowledgments} 

This work is supported by DARPA XAI project N66001-17-2-4029; ARO project W911NF1810296,  ONR MURI project N00014-16-1-2007, and Extreme Science and Engineering Discovery Environment (XSEDE) grant ASC170063. We thank Prof. Stu Geman, Prof. Xianfeng (David) Gu, Prof. Yali Amit,  Prof. Jun Zhang, and Prof. Chao Gao for helpful discussions.

\bibliography{mybibfile}

}

\end{document}